%% file: neurips2026/neurips_2026.tex
\documentclass{article}

\usepackage{hyperref}
\usepackage{url}
 \usepackage[preprint]{neurips_2026}

\usepackage[utf8]{inputenc} 
\usepackage[T1]{fontenc}    
\usepackage{hyperref}       
\usepackage{url}            
\usepackage{booktabs}       
\usepackage{amsfonts}       
\usepackage{nicefrac}       
\usepackage{microtype}      
\usepackage{xcolor}         
\usepackage{graphicx}
\usepackage{makecell} 
\usepackage{multirow}
\usepackage{wrapfig}
\usepackage{subcaption}
\usepackage{amsmath}
\usepackage[most]{tcolorbox}

\title{What Probing Reveals about Autonomous Driving:\\ Linking Internal Prediction Errors to Ego Planning}

\author{%
\begin{tabular}{c}
Hyeonchang Jeon$^{1}$ \quad
Kyungbeom Kim$^{1}$ \quad
Eugene Vinitsky$^{2,\ddagger}$ \quad
Kyung-Joong Kim$^{1,\ddagger}$ \\[0.4em]
$^{1}$Gwangju Institute of Science and Technology (GIST) \quad
$^{2}$New York University \\[0.3em]
\texttt{\{kevinjeon119, kyungbeom8\}@gm.gist.ac.kr} \\
\texttt{vinitsky.eugene@nyu.edu} \quad
\texttt{kjkim@gist.ac.kr}
\end{tabular}
}
\begin{document}

\maketitle
\begingroup
\renewcommand{\thefootnote}{\fnsymbol{footnote}}
\footnotetext[3]{Equal advising.}
\endgroup
\input{sec/0_abstract}

\input{sec/1_intro}

\input{sec/2_related_work}
\input{sec/3_experiment}
\input{sec/4_conclusion}


\bibliographystyle{plainnat}
\bibliography{neurips2026/ref}


\appendix
\input{}


\newpage
\input{sec/X_suppl}

\input{neurips2026/checklist}

\end{document}

%% file: sec/0_abstract.tex
\begin{abstract}
Large-scale datasets and fast simulators have enabled improvements in driving policies that appear safe and robust, yet strong performance in nominal scenarios can still mask flawed reasoning and unsafe heuristics. Summary scores from closed-loop simulators do not give significant insight into the policy, making it difficult to determine whether they truly predict the motion of surrounding vehicles, how the ego vehicle generates future plans, or whether they merely rely on brittle heuristics that happen to succeed in nominal scenarios. To better understand the limits and weaknesses of driving policies, we focus on probing for forms of \textit{prediction}, i.e., where surrounding vehicles will move next, and \textit{planning}, i.e., understanding how to generate safe trajectories.
We focus on these two capabilities because they reflect behaviors expected of effective driving policies, and use their presence or absence to assess policy quality across data-driven behavior cloning and simulation-driven reinforcement learning policies.  
To evaluate the presence of these capabilities, we investigate them as a function of scale, asking whether the closed-loop gains from larger datasets and longer simulation training reflect stronger prediction and planning or merely better behavioral heuristics. We use linear probing and targeted perturbations in both imitation learning and reinforcement learning models to track when these internal signals emerge, plateau, or fail.
Despite good closed-loop performance, policies often fail to form timely surrounding-vehicle predictions during near-collision events, revealing a limitation in the predictive signals available for ego planning.
Finally, causal intervention shows that correcting mistaken predictions improves ego planning toward safer trajectories.
\end{abstract}

%% file: sec/1_intro.tex
\section{Introduction}
\label{sec:intro}
The combination of new, large datasets \citep{ettinger2021large,caesar2020nuscenes,chang2019argoverse,wilson2argoverse} and GPU-accelerated simulation environments \citep{kazemkhani2025gpudrive,gulino2023waymax} has led to rapid advances in data-driven autonomous driving \citep{shi2022motion,nayakanti2023wayformer,sima2024drivelm,philiontrajeglish,feng2023trafficgen,hu2022model}. 
Imitation learning and reinforcement learning are core components of many of these advances, being used both for motion prediction, enabling the self-driving car to understand where objects in the scene will move, and for generating suggested trajectories or actions for the self-driving car itself.

However, the success of driving policies obscures several potential challenges. Although the available training data is relatively abundant, it remains far smaller than in areas such as NLP and is heavily biased toward nominal, uninteresting scenarios in which drivers simply drive straight~\citep{zhai2023rethinking,li2024ego,jeon2024beyond,makansi2021exposing}. Furthermore, scenarios that truly test the generalization of driving models are rare, leaving open whether these models learn generalizable rules or instead rely on spurious correlations and shortcut heuristics.\input{Figures/tex/main} These limitations and evaluation gaps raise a question: Are driving agents truly learning the core prediction and planning skills required for safe and robust driving \citep{makansi2021exposing,sun2024causalagents}?

There are several natural indicators we could use to assess understanding of driving policies; conversely, their absence would suggest a failure to internalize key skills. First, a skilled policy should respond sensibly to perturbations in simulation: removing a lead vehicle should allow the ego to accelerate, introducing one should induce deceleration, and reduced interaction should generally make driving easier, yielding increased driving progress and fewer collisions. Second, safe driving requires reasoning about surrounding vehicles, so a good policy would likely attend to a relevant subset rather than treating all vehicles equally or ignoring them. For example, it should respond to a slowing lead vehicle while discounting distant vehicles that do not affect the ego plan. Third, robust safety requires planning for contingencies, which likely entails maintaining internal predictions of safety-critical trajectories. For example, a truly skilled policy should be able to respond to abrupt cut-ins and maintain potential plans to handle unexpected scenarios.

From these requirements—reasoning about surrounding vehicles and maintaining an internal plan—we narrow our focus to two concrete, probeable competencies. We define \textit{prediction} as inferring the future positions of surrounding vehicles, and \textit{planning} as the adaptiveness of the ego autonomous vehicle (AV)'s planning in response to changes in predictions during decision-time planning \citep{bushinterpreting} as shown in Figure \ref{fig:main}. 
However, a driving model can perform well on driving metrics while failing to internalize these capabilities. Indeed, there is evidence that one can perform well on standard benchmarks while entirely neglecting any information but ego state~\citep{li2024ego} or achieve high-scoring performance while having incorrect causal models~\citep{sun2024causalagents}. Even if these failure modes are avoided, simple rules like “identify a lead vehicle and follow it at a safe distance" can be effective on most scenes while failing critically on the rare scenarios that make the rule unsafe. 

In this paper, we investigate the internals and behavior of driving models ranging from simple behavior cloning (BC) to state-of-the-art imitation learning (IL) and reinforcement learning (RL) policies, with the goal of understanding what aspects of safe driving they learn to represent. In particular, we study how surrounding-vehicle information is represented and used, and how prediction and adaptive planning emerge as data scale increases. We train driving models at multiple dataset sizes \citep{zheng2024preliminary,naumann2025data,baniodeh2025scaling} and evaluate each scale using perturbed closed-loop simulation, probing, and targeted interventions.

To test dependency on the presence of often irrelevant surrounding vehicles, we run \textit{perturbed simulations} in which we randomly remove them. To quantify what the model internally knows about nearby agents, we use \textit{surrounding-vehicle linear probing} \citep{alain2016understanding} to decode future positions of nearby agents from internal representations, and benchmark the quality of the internal representation against probes trained directly on raw inputs. We further analyze simulated near-collision events to examine how probing performance relates to actual collision outcomes.
Lastly, to study how surrounding-vehicle information influences planning, we use \textit{ego-linear probing} to decode the ego AV’s planned future positions and perform a decision-time \textit{intervention experiment} that modifies internal representations through these probes, testing whether correcting surrounding-vehicle mispredictions leads to safer and more appropriate ego plans \citep{bushinterpreting}.

From our results, can we say whether \textit{“better internal predictions translates to better plans?”} Often, yes: when provided with correct surrounding-vehicle predictions, the driving models reliably reroute or adjust their speed to produce safe and reasonable plans, and interventions that substitute correct predictions for incorrect ones steer trajectories back on course and reduce collisions in the internal representations. 
Our contributions are as follows:
\begin{itemize}
    \item \textbf{Prediction/planning probes across scales:} We develop linear probes to show that, as training data scale increases, models learn stronger planning representations and better ignore irrelevant vehicles, yet still struggle to identify truly safety-critical ones.
    \item \textbf{Perturbed closed-loop evaluation:} We introduce a surrounding-vehicle removal protocol and find that driving models perform worse even when driving should become easier in single AV driving, suggesting that they still partially rely on surrounding vehicles in a spurious way.
    \item \textbf{Near-collision analysis and causal intervention:} We use the probing to show that models attend to critical surrounding vehicles too late in near-collision events, and use interventions to show that ego planning can adapt with changed predictions and improve when mispredictions are corrected. 
\end{itemize}

%% file: Figures/tex/main.tex

\begin{figure*}[t]
    \vspace{-8pt}
    \centering  
    \includegraphics[width=0.95\linewidth]{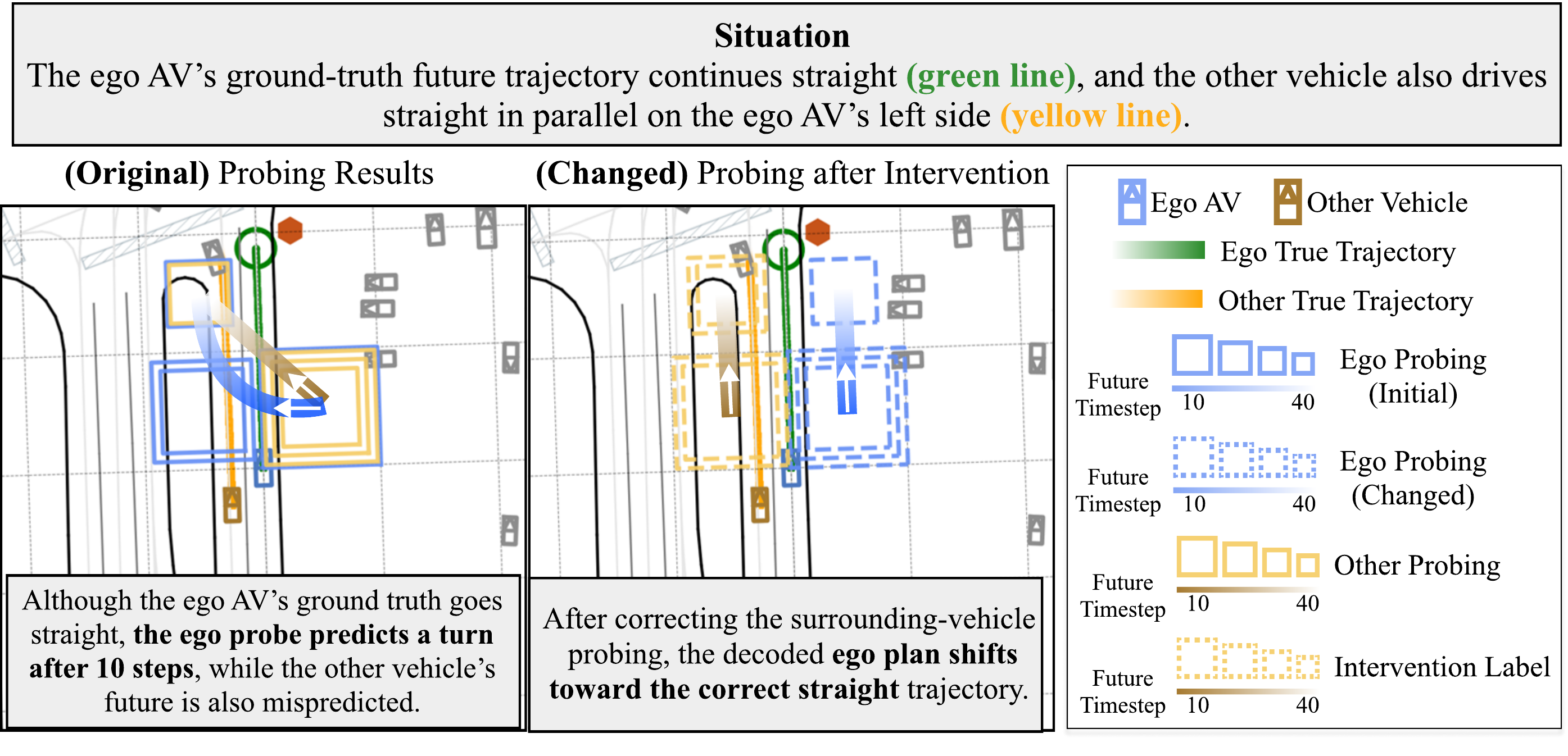}
    \caption{\textbf{Recovery of planning with correction of prediction.} The ego AV continues straight, while the other vehicle drives in parallel on the ego AV's left side. Gradient arrows simplify the probing results and show the decoded future motion direction over time. \textbf{(Original)}: After the 10-step prediction, the decoded ego plan curves away from its straight ground-truth trajectory, while the other vehicle's future is also mispredicted. \textbf{(Changed)}: Correcting the surrounding-vehicle prediction shifts the decoded ego plan toward the correct straight trajectory.}
    \vspace{-4mm}
    \label{fig:main}
\end{figure*}

%% file: sec/2_related_work.tex
\section{Related Work}
\subsection{Planning in Autonomous Driving}
Autonomous driving research has access to abundant datasets of human behavior \citep{caesar2020nuscenes,chang2019argoverse,ettinger2021large}. However, these datasets are heavily skewed toward nominal, low-interaction scenarios, leaving open the question of whether standard closed-loop simulation—typically evaluated on the same distribution—can truly measure robust behavior in safety-critical interactions. Standard planning models \citep{nayakanti2023wayformer,chai2020multipath,seff2023motionlm} are widely used for both prediction and planning, yet current evaluations rarely test whether they use surrounding-vehicle information in a timely and safety-relevant manner.

Recent studies \citep{li2024ego,zhai2023rethinking} have shown that source datasets are dominated by nominal straight-driving scenarios, in which the ego AV’s state alone often suffices to achieve strong closed-loop performance. As a result, closed-loop metrics can be misleading, indicating high performance even when the model largely ignores other vehicles. This motivates the need for alternative evaluations that directly assess whether models encode and use surrounding-vehicle information for prediction and planning. In this paper, we study how prediction and planning abilities in IL models are reflected in their internal representations and analyze perturbed simulation results using linear probing.

\subsection{Data Scaling Laws}
With the rise of foundation models, scaling laws have become an important research topic \citep{henighan2020scaling,tian2024visual,bharadhwaj2024roboagent}. They study how performance varies with dataset size, model capacity, and compute resources, especially for transformer-based models. This perspective has recently been extended to autonomous driving \citep{zheng2024preliminary,naumann2025data,baniodeh2025scaling}. For example, \citep{zheng2024preliminary} introduced ONE-Drive and showed that augmenting rare scenarios is more effective than randomly sampling scenarios. \citep{naumann2025data} studied end-to-end scaling through larger models and additional camera inputs. \citep{baniodeh2025scaling} found stronger alignment between open-loop and closed-loop metrics and identified compute-optimal model sizes. 
Similarly, we study how autonomous driving performance scales with dataset size and evaluate models with both metrics. Unlike prior work, however, we focus on how prediction and planning capabilities themselves scale with data.

\subsection{Interpretability in Deep Learning}
Interpretable deep learning aims to transform neural representations into forms that can be analyzed and linked to human-understandable concepts. Linear probing is a standard method for measuring what a trained network has learned from data using a probing dataset and a linear classifier \citep{alain2016understanding,mikolov2013linguistic}. It has been used to study planning abilities in both model-free RL \citep{kim2018interpretability} and supervised agents \citep{guez2019investigation}. In autonomous driving, prior work applies linear probing to frozen policy encoders trained from expert actions, using simple heads to evaluate affordance prediction and interpretable control \citep{xiao2021action}. More recent work \citep{tas2025words} explores probing-based methods with sparse autoencoders \cite{bricken2023towards} to disentangle latent factors and interpret or control ego motion. Our work takes a similar probing-based perspective, but focuses on multi-agent settings, where the probes target not only ego planning but also predictions about surrounding vehicles.

%% file: sec/3_experiment.tex
\section{Approaches}
In this section, we analyze how driving policies represent and use surrounding-vehicle information for decision making: (i) whether they can predict safety-relevant agents, (ii) whether such predictions are timely in near-collision situations, and (iii) how prediction representations causally influence planning via targeted interventions.
First, we show that both simple behavior cloning and RL models follow a similar scaling law to prior work \citep{baniodeh2025scaling,naumann2025data} (Section~\ref{sec3:scaling}). This serves as a sanity check that the models we probe exhibit the expected closed-loop improvements with scale, allowing us to ask whether these performance gains are reflected in their internal representations for prediction and planning.
Our probing results show that internal representations improve over raw inputs, but the gains are concentrated on simple behaviors. Distance-based analysis further shows that scaling helps models ignore distant vehicles, but not fully identify the most safety-relevant ones. (Section~\ref{sec3:lp-other}).
To understand how the presence of surrounding vehicles affects the driving model's performance, we conduct a perturbation simulation in which vehicles are randomly removed from the scene. We further relate closed-loop behavior to internal representations by analyzing how surrounding-vehicle probing evolves during a near-collision event (Section~\ref{sec3:ratio}).
Lastly, we ask: what happens if we correct these mispredictions? We show that, once aligned, the driving model can recover its planning and even exhibit collision-avoidance behavior (Section~\ref{sec3:lp-intervention}).

\begin{tcolorbox}[ colback=black!5, colframe=black!60, title=\textbf{Summary of Results.}, fonttitle=\bfseries, boxrule=0.6pt, arc=2pt, left=4pt, right=4pt, top=4pt, bottom=4pt ]
Scaling experiments with open-loop metrics and closed-loop simulation results improve overall driving performance, but they do not, by themselves, show whether models learn safety-relevant prediction and planning. Our probes show that driving models encode better planning and suppress clearly irrelevant surrounding vehicles, yet they still fail to emphasize truly safety-critical vehicles early enough in near-collision situations. Finally, intervention results suggest that more accurate internal predictions can causally improve ego planning, steering decoded plans toward safer and more appropriate trajectories.
\end{tcolorbox}

\subsection{Experimental Setup}
\label{sec3:setup}

\textbf{Models.}
We compare a transformer-based behavior cloning policy similar to \citet{gulino2023waymax} (BC) and a PPO-based RL policy \citep{schulman2017proximal,cornelisse2025building} (RL) across data scales. 
We additionally apply our linear probing and causal intervention analyses to SMART \citep{wu2024smart} (IL), a multi-agent trajectory prediction model, demonstrating that our probing-based framework extends to state-of-the-art IL models. 
Detailed model descriptions are provided in Appendix \ref{app:model_details}.

\textbf{Dataset.}
For the BC and RL models, we train our model using the Waymo Open Motion Dataset (WOMD)~\citep{ettinger2021large}. The dataset consists of over $400K$ driving scenes, each containing 9 seconds of trajectory data sampled at 10 Hz, with up to 128 cars per scene. Due to limitations of the underlying simulator used to replay the trajectories, we exclude scenes containing traffic lights and overpasses, filtering the dataset to $\sim 90K$ scenes. Each observation is partially observable and includes information about the ego vehicle, surrounding vehicles (up to 127), and the map in an ego-centric view. As action labels are not available in the dataset, we derive them through inverse kinematics. 
For the IL model, we set the same setting in SMART \citep{wu2024smart}. We train using the WOMD dataset with 11 historical steps as input, including global coordinate, heading information, and predict the remaining future trajectories for the target. Since the IL model predicts multi-agent trajectories, it can predict up to 32 vehicles per scenario.

More details on the dataset can be found in Appendix \ref{app:dataset}. For a detailed analysis of closed-loop simulations, we first preprocess trajectories to remove those with infeasible kinematics (likely due to noise during data collection), and then classify the remaining trajectories into mutually exclusive categories (\textit{Straight}, \textit{Turn}, \textit{Reverse}, and \textit{Uncategorized}). (See details in Appendix \ref{app:trajectory_types})

\textbf{Linear Probing.}
To interpret the driving model's internal representation, we perform a linear probing experiment. We select the best model across seeds for each dataset scale and train a linear classifier to predict future positions.
The ground truth is set as the vehicle’s positions (ego or surrounding vehicle) 1 to 4 seconds ahead (corresponding to 10, 20, 30, and 40 steps). The positions are discretized into 64 labels by dividing the ego vehicle’s current field of view into an $8\times8$ grid along the x- and y-axes, normalized with respect to the ego AV’s current position. 
To understand what information is retained in the internal network layers, we train linear classifiers on both the raw input and representations from the early and late attention layers. We evaluate performance using the F1 score to account for data imbalance, as well as per-trajectory-type accuracy to investigate how the driving models learn differently across cases. 
For the IL model, which predicts multi-agent trajectories, there is no predefined ego agent. We therefore sample one vehicle as the ego and select its nearest neighboring vehicle as the surrounding vehicle. Since our goal is to analyze how information about surrounding vehicles contributes to ego-trajectory prediction, this setup provides a natural egocentric perspective for evaluating the use of information from other agents.
The linear probing setting is described in Appendix \ref{app:lp-other-setting} and Appendix \ref{app:lp-ego-setting}.

\subsection{Data Scaling Laws}
\label{sec3:scaling}
\input{Figures/tex/data_scale}
To investigate the emergence of planning and prediction ability, we train the BC and RL models with three different random seeds while gradually increasing the dataset size, and evaluate them in the GPUDrive simulator \citep{kazemkhani2025gpudrive} on unseen scenarios. 
For RL, we train on up to 10K scenes, at which point performance has nearly converged, as shown in prior work \citep{cornelisse2025building}.
To support our model's generality, we evaluate our model performance using the Waymo Open Sim Agents Challenge (WOSAC) metrics \citep{montali2023waymo}.
For closed-loop simulation, we use three metrics— vehicle collision (denoted Veh-Coll), the off-road rate, and goal progress ratio which is calculated by $ 1 - d_{final} / d_{initial}$ where $d_{final}$ denotes the distance to the goal at the final timestep and $d_{initial}$ denotes the distance to the goal at the beginning of the trajectory. 

As shown in Figure \ref{fig:scaling-law}, metrics decrease with scale, following a curved power law that approaches a plateau after $\approx20K$ scenes for the BC model, which has shown similar trends with \citep{baniodeh2025scaling}. In the case of the RL model, it achieves much lower overall collision rates (Veh-Coll and Off-Road) while surpassing the goal progress after 1K scenes, ultimately reaching 99\% of the goal progress ratio. Both BC and RL models achieve high correlation, indicating that scaling laws hold in closed-loop simulation.
See Appendix \ref{app:train-validation} and Appendix \ref{app:scale-law} for more results.

\subsection{Linear Probing for Surrounding Vehicles Prediction}
\label{sec3:lp-other}
\input{Figures/tex/other_probe}

To understand how the driving model utilizes surrounding-vehicle information internally, we train linear probes on the intermediate representations of our trained model to evaluate its prediction and planning capabilities, employing the linear probing method \citep{alain2016understanding}. We train the linear probes with three different random seeds while gradually increasing the dataset size.
Since the models differ in architecture and input dimensionality (i.e., the RL model uses a single timestep, whereas the BC model uses five timesteps), directly comparing raw probing performance across models can be misleading. Therefore, for cross-model comparison, we report the relative probing gain, computed as the difference between the trained model's probe and the raw-input probe (LP $-$ Raw). This measures how much additional information is encoded in the learned representation beyond what is easily available from the raw input.

As shown in Figure \ref{fig:other-probing}, in the case of the BC model, the F1 Macro score difference is increasing until 10K scenes and becomes plateaued. This trend is similar to the closed-loop simulation results in Section \ref{sec3:scaling}, where surrounding-vehicle probing performance nearly peaks at 10K scenes, then slightly decreases before plateauing. Interestingly, the RL model initially maintains a higher F1 score than BC across all timesteps, but this advantage diminishes substantially by 10K scenes, though it still remains above BC. As shown in Appendix \ref{app:detail-ego-probing}, we find that the RL model gradually places greater emphasis on ego planning. Across timesteps, BC tends to show lower F1 scores as the prediction horizon increases, whereas RL largely maintains its F1 difference over time. Lastly, while SMART (IL) is comparable to 80K BC at the 10-step horizon, it shows a clearer advantage at longer horizons. In particular, its LP--Raw difference remains stable or slightly increases from 10 to 40 steps, whereas BC tends to decline. This suggests that SMART (IL) better preserves information about surrounding vehicles over longer temporal ranges.

For type accuracy, BC continues to improve with scale, and this improvement is particularly pronounced for the relatively easier $Uncategorized$ and $Straight$ cases, while the gap remains much smaller for $Reverse$ and $Turn$ cases. In contrast, RL shows a largely similar pattern across scales, with little noticeable change. IL shows relatively stable type-accuracy differences; unlike its F1 Macro trend, it does not clearly outperform BC at the type level. This suggests that IL has an advantage in long-horizon planning, while BC and RL, which directly output actions, tend to focus more on shorter-horizon signals. More detailed results are provided in Appendix \ref{app:other-probing}.

\input{Figures/tex/other_analysis}
However, this nominal understanding does not directly assess predictive ability, since only a few vehicles are actually important among the many surrounding vehicles in each scenario. To address this, we further analyze whether the surrounding-vehicle linear probe is stronger for vehicles that will become closer to the ego in the future. In Figure \ref{fig:other-analysis}, we plot the correlation between linear-probe prediction and future ego–other distance on both BC and RL models. The vertical axis shows the difference in predicted probability for the true label between the LP and the raw-input probe, while the horizontal axis shows the future distance 10 steps ahead. Points are colored by relative distance change, $\frac{current - future}{current}$: samples that become more than 40\% closer are shown in red, whereas those that become more than 40\% farther are shown in blue, with opacity increasing with the magnitude of the change.

At a small scale (100 scenes), the linear-probe advantage over the raw-input probe is only weakly associated with future ego--other distance, indicating limited sensitivity to future interaction relevance. With more training data (10,000 and 80,000 scenes), the association becomes markedly more negative, indicating that the representation increasingly favors vehicles that will remain close or move closer to the target. However, among nearby vehicles, the model still attends to both approaching and receding vehicles. This suggests that scale helps the model filter out clearly irrelevant vehicles, but it does not yet fully isolate or predict the truly critical vehicles, i.e., those that will move even closer to the target.

\subsection{Randomly Removing Surrounding Vehicles}
\label{sec3:ratio}
\input{Figures/tex/perturbed}
If the driving model does not spuriously depend on the  presence of surrounding vehicles, removing them should not degrade performance and may even improve goal-reaching performance by reducing interaction constraints.
Moreover, stronger prediction ability should lead to more robust planning across varying numbers of surrounding vehicles. To evaluate this, we randomly remove a fraction of active vehicles from each of 10,000 validation scenes, as shown in Figure \ref{fig:perturbed}.
In both models, performance drops in the single-AV setting, with increased off-road rates and worse goal-progress trends. 
This is undesirable: removing surrounding vehicles should make the driving easier, not harder. The degradation, therefore, suggests that the policies partially rely on the presence of the other vehicles in a spurious way, rather than using them as an interaction.
This likely reflects WOMD’s bias toward multi-vehicle trajectories. The drop decreases as data scale increases, suggesting that larger datasets partially mitigate this issue. See additional results in Appendix \ref{app:perturbed}.

\input{Figures/tex/near_collision}

Beyond the single AV setting, the perturbed simulation offers only a rough, indirect view of how the model handles surrounding vehicles. To more explicitly examine whether the model predicts the behavior of surrounding vehicles in safety-critical situations, we therefore conduct a near-collision analysis using the surrounding-vehicle linear probe.
We define a pre-collision window $W$ of 10 steps and compare the probing signal within this window against the model's average probing signal over the full validation episodes. For each spatial grid cell $grid$, we compute the normalized difference $(w-a_grid)/a_grid$, where $a_grid$ is the corresponding full-episode average. We filter out grid cells with fewer than 100 collision cases and mark them in dark gray.

Figure~\ref{fig:collision} visualizes this normalized difference at 40, 30, 20, and 10 steps before collision.
Red grids denote spatial regions whose surrounding-vehicle probing signal increases relative to the whole episode average, whereas blue grids denote regions whose signal decreases. Therefore, early anticipation of a dangerous interaction would appear as a strong positive signal around the relevant nearby vehicle at earlier horizons, such as 40, 30, or 20 steps before the event. In contrast, for the BC model, the signal remains weak and diffuse at earlier horizons and becomes clearly concentrated near the ego only in the final 10 steps before collision. Although this late increase indicates that the model eventually represents the nearby colliding vehicle, it is already too late for effective avoidance; therefore, these episodes remain planning failures.

The RL model shows a similar qualitative pattern, but collision cases are much rarer, resulting in substantially fewer samples and larger dark-gray filtered regions. Despite this sparsity, the strongest positive signal again appears mainly in the final 10-step window. Overall, both BC and RL appear to use surrounding-vehicle information near a collision, but they fail to emphasize the relevant vehicle early enough to avoid the crash.
We analyze near-collision events for off-road in Appendix \ref{app:near-collision}.

\subsection{Intervention Test for Adaptiveness of Ego AV Planning}
\label{sec3:lp-intervention}
What happens to the model's planning when it has more accurate predictions, and how does it adapt to changes in the predictions of surrounding vehicles?
To test this, we first applied linear probing to the ego AV’s planning, as in the Section \ref{sec3:lp-other}, which we refer to as ego-linear probing. The result of ego-linear probing is in Appendix \ref{app:ego-probing}.

Inspired by \citet{bushinterpreting}, we intervene on internal representations to test whether ego planning causally adapts to perturbed surrounding-vehicle predictions and recovers when incorrect predictions are corrected.
We modify the earlier layer representation of a selected surrounding vehicle $o$ and examine how this perturbation propagates to the ego’s later layer probing.
Let the earlier layer feature be $g = [g_o;\, g_{-o}]$, $g_o \in \mathbb{R}^{128}$, the feature of surrounding vehicle $o$, and $g_{-o} \in \mathbb{R}^{127 \times 128}$ those of the remaining vehicles including ego vehicle. 
We obtain the surrounding-linear probing weight $w_l \in \mathbb{R}^{128}$ corresponding to the surrounding vehicles' future position label $l$ from a linear probe trained on earlier layer features of surrounding vehicles at future timestep $s\in S$ where $|S|$ is the number of future timesteps. Our intervention adds this label direction to $g_o$:
\begin{equation}
g_o' \;=\; g_o \;+\; \frac{1}{|S|}\alpha\sum_s w_l^s, 
\qquad
g' \;=\; [g_o';\, g_{-o}].
\label{eq:intervention}
\end{equation}
If the model has adaptive planning capability, the later layer activation $h' = f(g')$ should change coherently from the baseline $h = f(g)$ along a semantically meaningful direction $w_l$, scaled by a strength parameter $\alpha$. This encourages a change in the ego-linear probing output, from $p$ to $p'$, as predicted by the linear probe layer $e$.
\begin{equation}
h' \;=\; f(g'),
\qquad
p' = e(h').
\label{eq:intervention2}
\end{equation}

\input{Figures/tex/intervention}

we conduct intervention experiments in two settings: \textit{adaptiveness} and \textit{recovery}. Adaptiveness measures whether the ego plan changes in response to a potential collision. To test this, we perturb the surrounding-vehicle probing representation to overlap with the ego representation and observe the resulting change in ego planning. Recovery instead examines whether correcting an incorrect prediction of the surrounding vehicle restores a more appropriate ego plan.

For each model, we label 100 validation scenes and remove irrelevant cases such as short trajectories or single-AV scenes. This leaves 59 valid cases for BC (43 adaptiveness, 16 recovery), 58 for RL (53 adaptiveness, 5 recovery), and 64 for IL (51 adaptiveness, 13 recovery). 
We summarize the overall intervention results across all models in Appendix \ref{app:intervention}. 
Overall, both models frequently adapt their plans when predictions about surrounding vehicles are updated, and recovery interventions often restore more appropriate planning. 
The results also reveal model-specific tendencies: BC is more reliable for route-change and slowdown interventions but struggles with speed-up cases, whereas RL shows a more balanced response across intervention types, and the IL shows good at adjusting speed both slower and faster. (See Table \ref{tab:intervention-label} in Appendix \ref{app:intervention} for all results)

Figure \ref{fig:intervention} shows that the ego's planning can shift from an initially unsafe trajectory to a safer one that avoids collisions with surrounding vehicles. Moreover, in failure cases, restoring the incorrect predictions of surrounding vehicles led the models to reorient their planning toward the goal, indicating that accurate predictions help generate correct plans.
This effect appears across all three models: across both intervention types, the ego-planning probe changes in 36 of 59 BC cases, 37 of 58 RL cases, and 40 of 64 IL cases, suggesting that surrounding-vehicle predictions causally affect ego planning.
However, the ego plan sometimes remains roughly aligned with the ground-truth trajectory even with imperfect predictions of surrounding vehicles, suggesting that precise predictions are not always necessary for effective planning. (see Appendix \ref{app:intervention} for more example cases).

%% file: Figures/tex/data_scale.tex
\begin{figure*}[h]
    \vspace{-8pt}
    \centering  
    \includegraphics[width=0.95\linewidth]{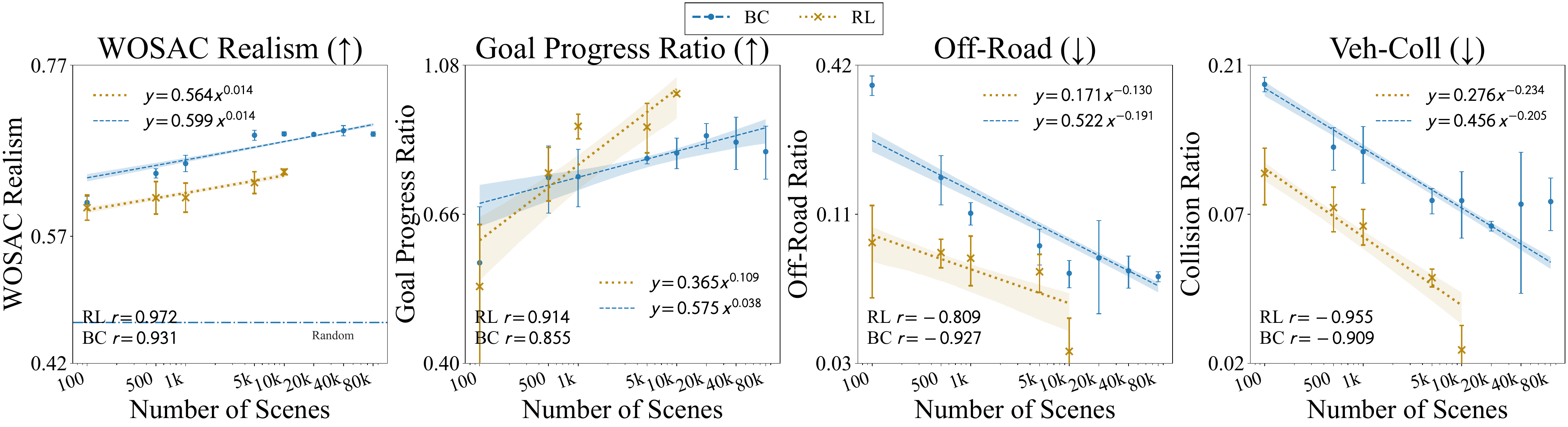}
    \caption{\textbf{Power law relationships of BC and RL models}: We evaluate the WOSAC, collision metrics (Off-Road and Veh-Coll), and goal progress rate. $r$ is the correlation coefficient.}
    \vspace{-3mm}
    \label{fig:scaling-law}
\end{figure*}

%% file: Figures/tex/other_probe.tex
\begin{figure*}[h]
    \vspace{-3mm}
    \centering
    \includegraphics[width=0.97\linewidth]{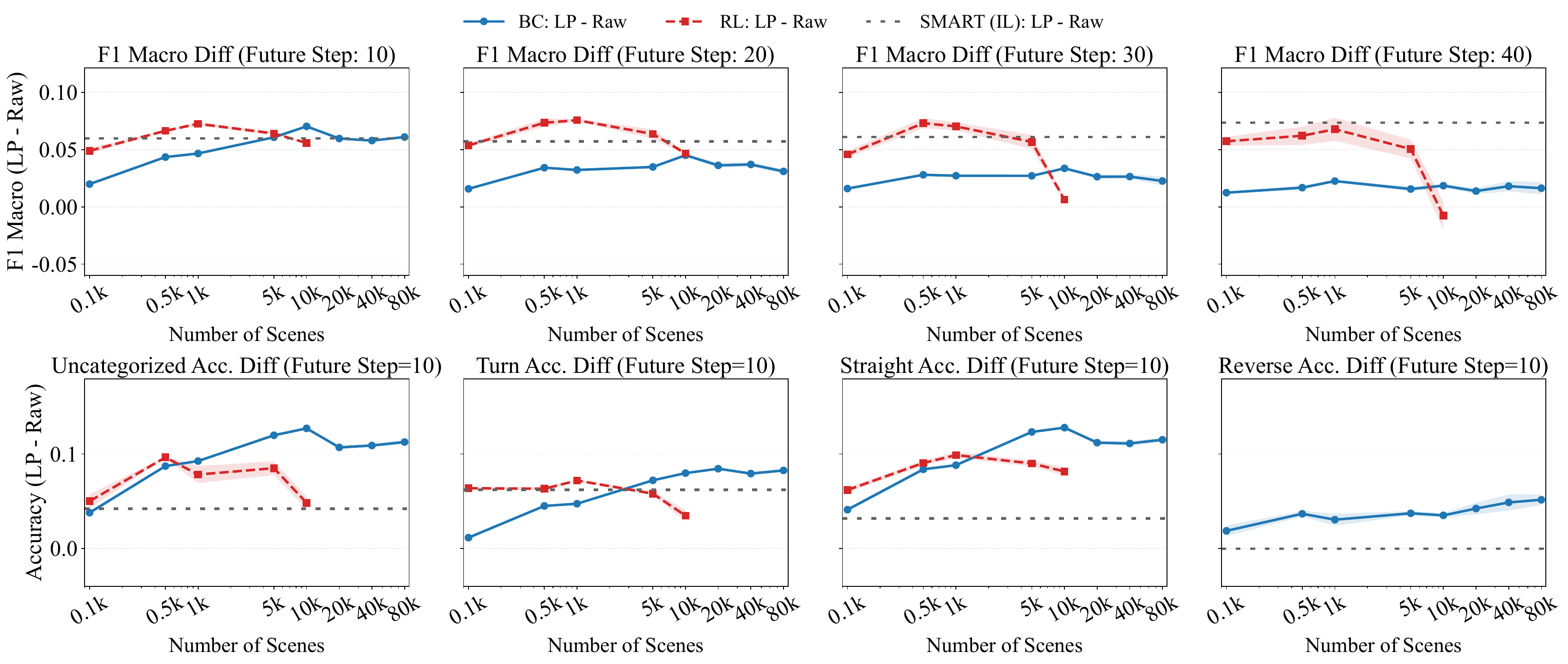}
    \caption{\textbf{Performance metrics with surrounding-vehicle linear probing.}
    Top: F1 Macro score difference between linear probes on trained model representations and the raw-input baseline (LP $-$ Raw) across future steps $fs \in {10,20,30,40}$, where $fs$ denotes the prediction horizon in timesteps.
    Bottom: trajectory-type accuracy difference at $fs=10$ for $Straight$, $Turn$, $Reverse$, and $Uncategorized$ cases.
    Higher values are better for both metrics.
    The $Reverse$ case of RL is omitted because no $Reverse$ cases are available for the RL setting.}
    \label{fig:other-probing}
\end{figure*}

%% file: Figures/tex/other_analysis.tex
\begin{figure*}[h]
    \centering  
    \includegraphics[width=0.97\linewidth]{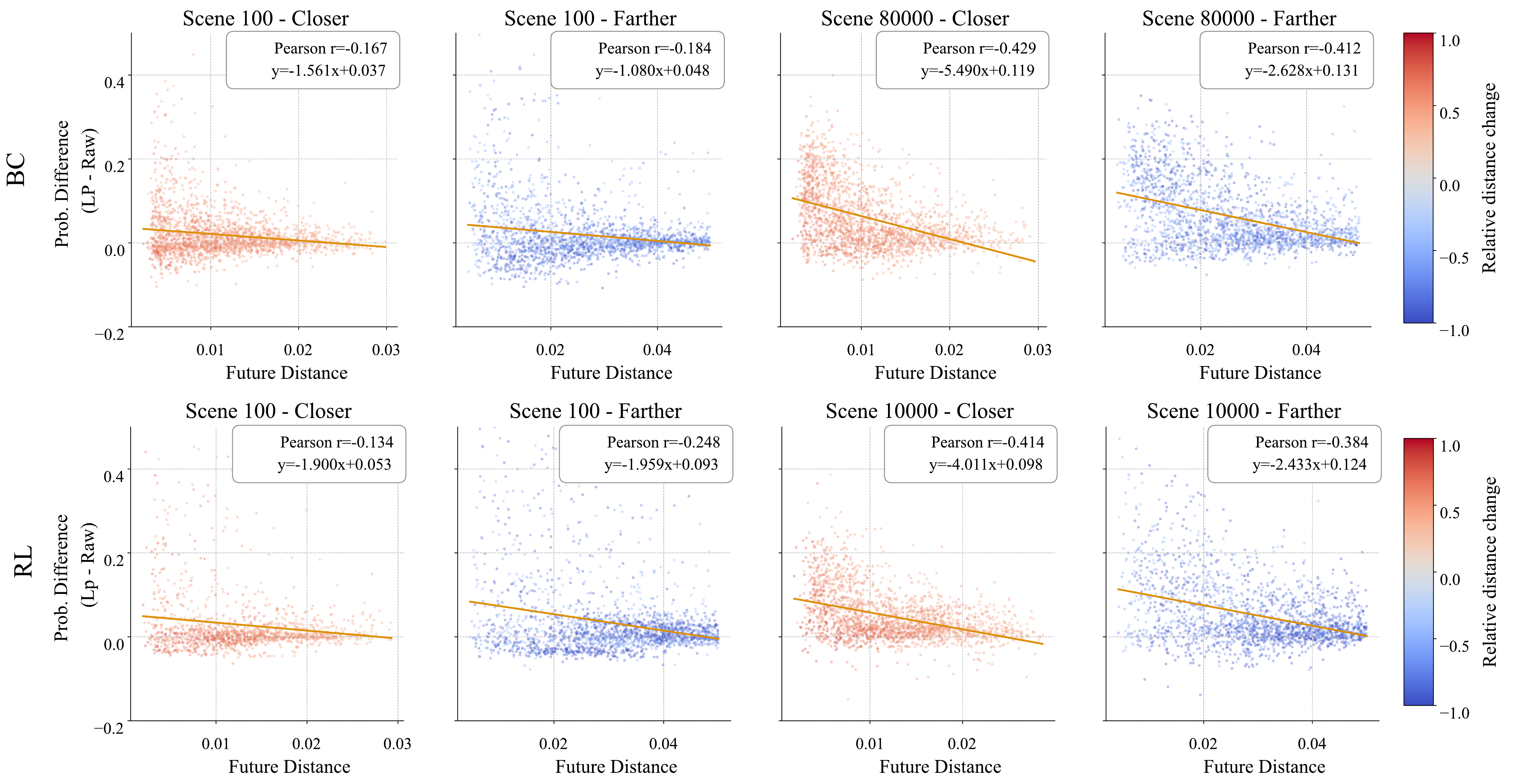}
    \caption{\textbf{Correlation between linear-probe and future ego--other distance.}
    Vertical: probability gap (LP -- raw); horizontal: future ego--other distance.
    Orange: regression line; $r$: correlation.}
    \label{fig:other-analysis}
    \vspace{-1mm}
\end{figure*}

%% file: Figures/tex/perturbed.tex
\begin{figure*}[h]
    \vspace{-12mm}
    \centering  
    \includegraphics[width=0.94\linewidth]{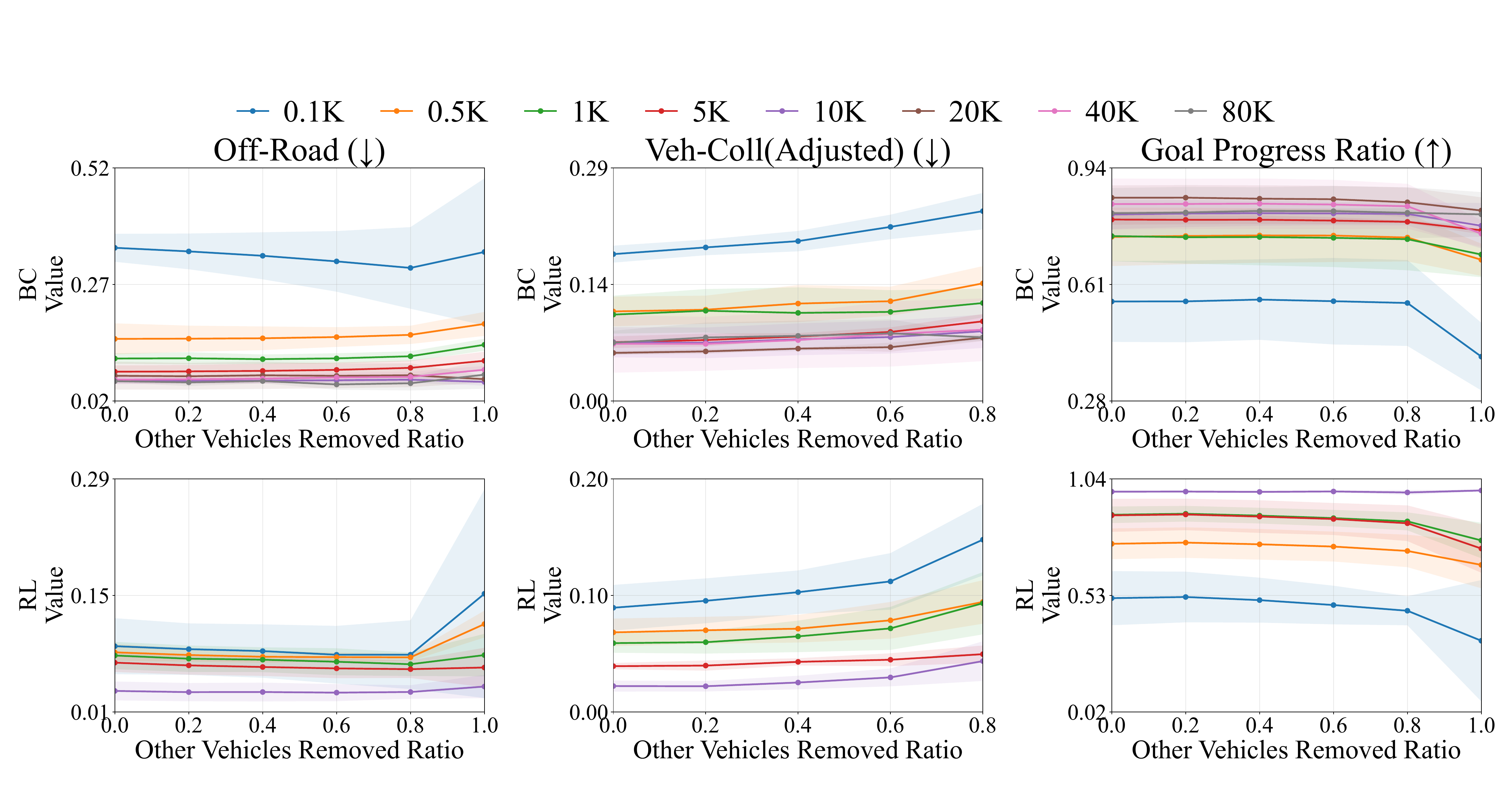}
    \vspace{-12pt}
    \caption{\textbf{Perturbed simulation results}: We test the IL model by randomly removing the surrounding vehicles with a ratio $p$. We adjust the ratio by multiplying $\frac{1}{ratio}$ for vehicle collision.}
    \label{fig:perturbed}
\end{figure*}

%% file: Figures/tex/near_collision.tex
\begin{figure*}[h]
    \centering
    \includegraphics[width=0.97\linewidth]{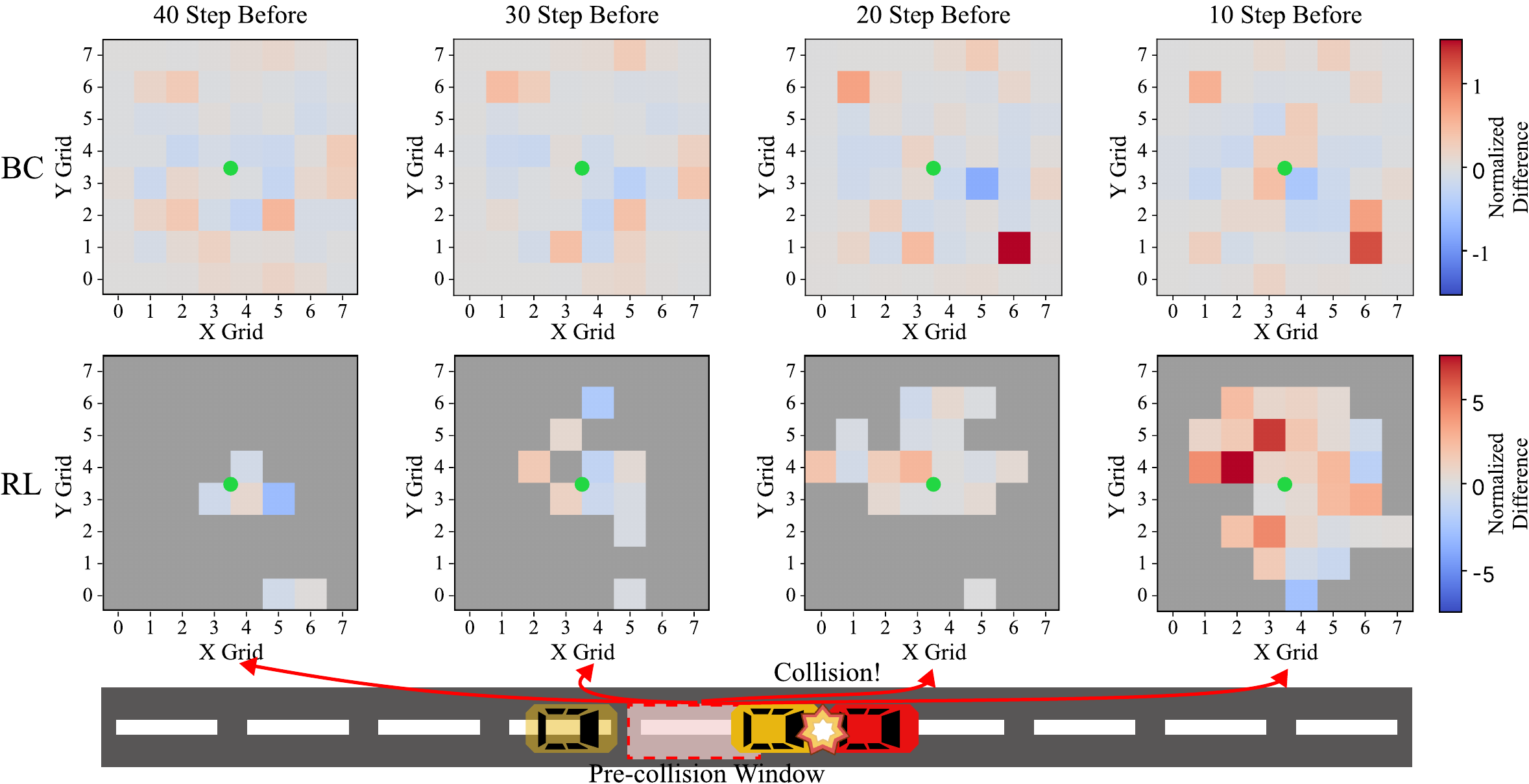}
    \caption{\textbf{Near-collision Analysis.}
    Each heatmap shows the normalized difference between the predicted probability of surrounding-vehicle probing in the pre-collision window $w$ and that of each spatial grid over the full episode, computed as $\frac{w-a_{\mathrm{grid}}}{a}$, at 10 to 40 steps before collision.
    The heatmaps use an ego-centric $8\times8$ grid covering $[-50,50]$ meter along both axes, so each grid cell corresponds to a $12.5\text{ m}\times12.5\text{ m}$ spatial region.
    Red grids mark areas that become increasingly important to the model near collision, whereas blue grids mark areas that become relatively less important compared with the model’s average attention over the full episode.
    The green dot marks the ego vehicle position. Dark gray cells indicate grids that were filtered out due to insufficient collision samples. Top: BC, Bottom: RL}
    \label{fig:collision}
    \vspace{-3mm}
\end{figure*}

%% file: Figures/tex/intervention.tex
\begin{figure*}[h]
    \centering
    \includegraphics[width=0.97\linewidth]{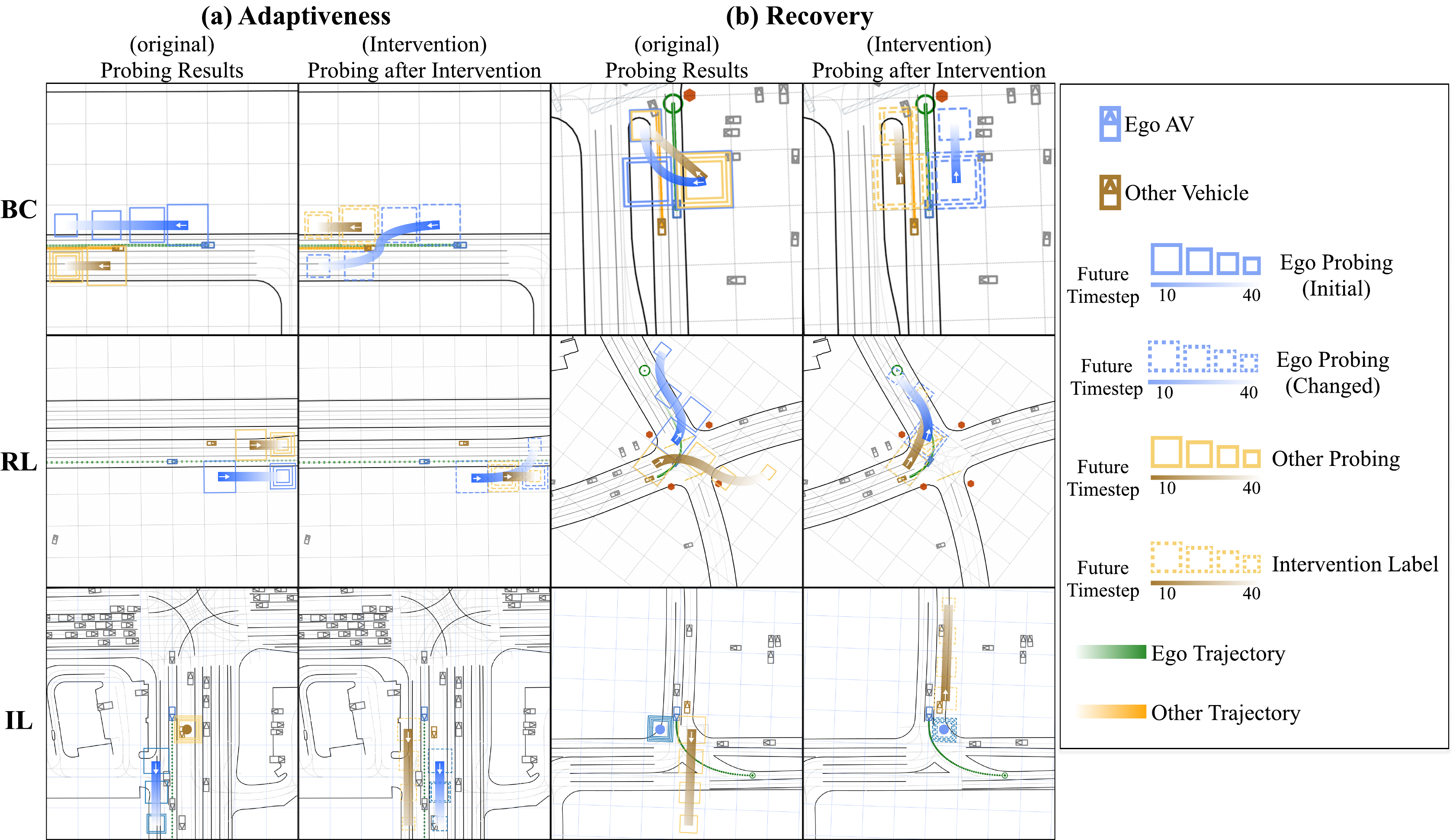}
    \caption{\textbf{Intervention experiment for adaptiveness and recovery (10 to 40 future timesteps)}: Examples from BC, RL, and IL models, and each pair of columns compares the original probing result with the probing result after intervention. (a) \textbf{Adaptiveness}: we perturb the surrounding-vehicle prediction so that it overlaps with the ego's predicted path, and test whether the ego plan changes to avoid the induced conflict. (b) \textbf{Recovery}: we replace an incorrect surrounding-vehicle probe with the correct intervention label, and test whether the ego plan is restored toward a safer or more ground-truth-aligned trajectory.}
    \label{fig:intervention}
    \vspace{-3mm}
\end{figure*}

%% file: sec/4_conclusion.tex
\section{Conclusion}
In this paper, we investigate how driving models preserve and use surrounding-vehicle information for ego planning. As the dataset size increases, the models become better at ignoring irrelevant agents. 
Furthermore, our perturbed closed-loop evaluation reveals that the reliance on surrounding vehicles is not always robust: removing all surrounding vehicles degrades the performance, even though the task should become easier in their absence. This suggests that the policies may partially rely on surrounding vehicles in a spurious way.
In particular, during near-collision events, it should anticipate others’ positions earlier, yet it often fails to do so and remains biased toward “nominal” vehicles, such as those going straight. Finally, we find that when prediction capability is restored, the planner produces stronger trajectories, suggesting that better prediction can lead to better planning. Moreover, the models not only recover but also actively avoid encroaching vehicles, indicating that their planning is already sufficiently robust to support collision avoidance.

These findings suggest that better planning may require not only stronger prediction but also more precise approaches for measuring the surrounding-vehicle information that the models capture.
Currently, our probing approach utilizes discretized predictions, which can be coarse and potentially blur fine-grained behavior. Designing probes and visualizations in continuous space is a promising direction. 
The correlation between surrounding-vehicle probing and future distance suggests that both BC and RL models can suppress some irrelevant agents. However, it remains unclear whether it reliably identifies the agents most critical for safe planning. To address this, incorporating joint future prediction \citep{luo2023jfp} or auxiliary tasks \citep{li2024interactive} to improve the model architecture, as well as providing additional language input signals from datasets \citep{malla2023drama,liwomd,chang2025langtraj}, may help the model develop a richer understanding of surrounding vehicles.

%% file: sec/X_suppl.tex
\input{appendix/a_model}
\input{appendix/b_dataset}
\input{appendix/c_scaling}
\input{appendix/d_surrounding}
\input{appendix/e_ego}
\input{appendix/f_further}
\input{appendix/g_intervention}

%% file: appendix/a_model.tex
\section{Model and Training Details}
\label{app:model_details}
\subsection{BC Details}
\begin{figure}[th]
    \centering
    \includegraphics[width=1.0\linewidth]{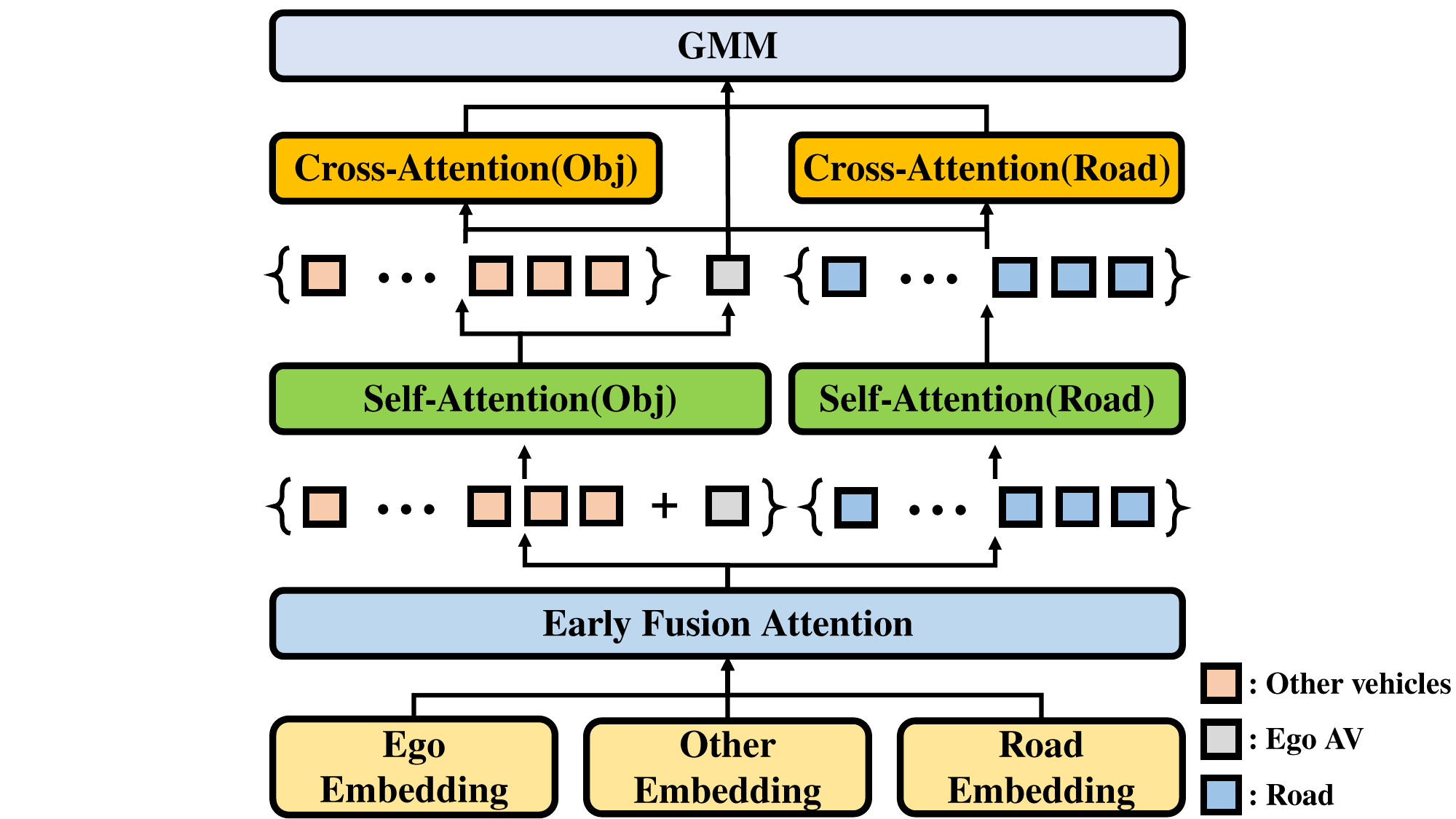}
    \caption{\textbf{Behavior Cloning model architecture}: Overall architecture of the behavior cloning model. Ego vehicle, other vehicles, and road features are first embedded and fused via early-fusion attention. The fused representations are refined via self- and cross-attention modules and finally modeled with a Gaussian Mixture Model (GMM).}
\end{figure}
In our behavior cloning framework, the model is conditioned on three types of inputs: 
ego vehicle state, other vehicles' state, and road context features in Table~\ref{tab:obs_features}.
Each input modality is first embedded into a latent representation through dedicated encoders:
\begin{align}
h^{ego}   &= f_{ego}(x^{ego}), \\
h^{other} &= f_{other}(x^{other}), \\
h^{road}  &= f_{road}(x^{road})
\end{align}
where $f_{ego}, f_{other}, f_{road}$ denotes a scene encoder applied to the corresponding features, 
implemented as a multi-layer perceptron with layer normalization and non-linear activations:
\begin{align}
z_1 &= W x + b, \quad && \text{(Linear)} \\
z_2 &= \text{Dropout}(z_1), \quad && \text{(Dropout)} \\
z_3 &= \text{LN}(z_2), \quad && \text{(Layer Normalization)} \\
h   &= \phi(z_3), \quad && \text{(Activation, e.g., Tanh)}
\end{align}
The embedded representations are first fused through an early fusion attention module, 
which enables the ego state to jointly attend to both road and object-level features:
\begin{equation}
H^{ego},H^{other},H^{road} = \text{Attn}([h^{ego}, h^{other}, h^{road}]).
\end{equation}
To refine the fused representation in a modality-specific manner, 
self-attention layers are applied independently to object-level and road-level embeddings:
\begin{align}
\tilde{H}^{ego},\tilde{H}^{other} = \text{SelfAttn}([H^{ego},H^{other}]), \\
\tilde{H}^{road} = \text{SelfAttn}(H^{road}).
\end{align}
To enhance modality-specific interaction and enable more informed policy learning, cross-attention layers are applied, allowing the ego representation to selectively attend to road-level and object-level features:
\begin{align}
\tilde{H}^{ego}_{other} = \text{CrossAttn}([\tilde{H}^{ego}, \tilde{H}^{other}]), \\
\tilde{H}^{ego}_{road} = \text{CrossAttn}([\tilde{H}^{ego}, \tilde{H}^{road}]).
\end{align}
Finally, the aggregated representation—obtained by concatenating the refined ego embedding 
with the outputs of the road- and object-conditioned cross-attention modules 
$(\tilde{H}^{ego}, \tilde{H}^{ego}_{other}, \tilde{H}^{ego}_{road})$ is fed into an output head parameterizing a Gaussian Mixture Model (GMM):
\begin{equation}
p(y \mid x) = \sum_{k=1}^{K} \pi_k \, \mathcal{N}(y \mid \mu_k, \Sigma_k),
\end{equation}
where $y$ denotes the predicted future action of the ego vehicle, and $\pi_k, \mu_k, \Sigma_k$ 
are the mixture weights, means, and covariances of the $k$-th Gaussian component, respectively. 
This probabilistic formulation enables the model to capture multimodal action distributions 
and inherent uncertainties in motion prediction.
During training, the model parameters are optimized by minimizing the negative 
log-likelihood (NLL) of expert demonstrations under the predicted GMM distribution. 
Specifically, given an expert action $y^{*}$, the likelihood under the mixture model is
\begin{equation}
p(y^{*} \mid x) = \sum_{k=1}^{K} \pi_k \, \mathcal{N}(y^{*} \mid \mu_k, \Sigma_k),
\end{equation}
and the loss function is defined as
\begin{equation}
\mathcal{L}_{\text{GMM}} = - \mathbb{E}_{(x,y^{*}) \sim \mathcal{D}} 
\left[ \log p(y^{*} \mid x) \right],
\end{equation}
where $\mathcal{D}$ denotes the dataset of expert demonstrations. 
This objective encourages the model to assign high probability density 
to expert actions, thereby aligning the predicted distribution with the 
expert policy.

\subsection{BC Training Settings}
Across all experiments of dataset size, the BC model has 1.4M parameters and is trained with a batch size of 512, a hidden layer size of 128, and a 6-component GMM head. We train with a learning rate of $0.0005$ and apply a weight decay. The different settings across different dataset sizes are in Table \ref{tab:il-setting}. We used 4 A100 GPUs to train BC models.
\label{app:train-setting}
\input{Tables/3_data_scaling_hyperparameters}

\subsection{RL Training Settings}
Across all experiments, we train a feed-forward IPPO policy \citep{schulman2017proximal,cornelisse2025building} with 1.2M parameters and a hidden dimension of 128. Training is performed with a batch size of 65,536 and a minibatch size of 4,096 for 2 update epochs, using a learning rate of 0.0003. We set the discount factor to 0.99, the GAE to 0.95, the clipping coefficient to 0.2, the entropy coefficient to 0.0001, and the value loss coefficient to 0.3. We scale the number of training steps with dataset size, using 0.1B, 0.2B, 0.3B, 0.75B, and 1B steps for 100, 500, 1,000, 5,000, and 10,000 scenes, respectively. We used 4 A100 GPUs to train RL models.
\newpage

%% file: Tables/3_data_scaling_hyperparameters.tex
\begin{table}[h]
\centering
\caption{IL Training Details.}
\label{tab:il-setting}

\begin{subtable}[t]{0.48\columnwidth}
\centering
\caption{Small scales}
\resizebox{\linewidth}{!}{%
\begin{tabular}{c|c|c|c}
\hline
\textbf{\#Scenes} & \textbf{\#Vehicles} & \textbf{\#Samples} & \textbf{Grad. Steps} \\
\hline
100   & 0.7K & 44K  & 20,000 \\
500   & 3K   & 215K & 50,000 \\
1000  & 7K   & 431K & 100,000 \\
5000  & 33K  & 2.1M & 250,000 \\
\hline
\end{tabular}}
\end{subtable}
\hfill
\begin{subtable}[t]{0.48\columnwidth}
\centering
\caption{Large scales}
\resizebox{\linewidth}{!}{%
\begin{tabular}{c|c|c|c}
\hline
\textbf{\#Scenes} & \textbf{\#Vehicles} & \textbf{\#Samples} & \textbf{Grad. Steps} \\
\hline
10000 & 68K  & 4.2M  & 400,000 \\
40000 & 270K & 16.9M & 700,000 \\
80000 & 542K & 33.8M & 850,000 \\
\hline
\end{tabular}}
\end{subtable}

\end{table}

%% file: appendix/b_dataset.tex
\section{Dataset Details}
\label{app:dataset}

\subsection{Observation Features}
The observation space provides a multi-modal representation composed of four key features: 
\textit{ego state features}, describing the ego vehicle’s kinematics and goal-related information; 
\textit{surrounding vehicle features}, encoding surrounding dynamic surrounding vehicles; 
\textit{road graph features}, capturing static road topology and structural elements (road edge, road line, road lane, crosswalk, speed bump, stop sign, and None as padding). For all experiments, we set the maximum number of agents per scenario to 128, and we consider the nearest 200 road points. 
The detailed specifications of these observation features, including their type, constituent variables, dimensionality, and description, 
are summarized in Table~\ref{tab:obs_features}.

\input{Tables/2_obs_features}
\subsection{Inverse Kinematics Model}
\subsubsection{Delta Dynamics Model}
\label{app:delta-local}
In our simulation environment, we employ a delta dynamics model \citep{gulino2023waymax} to update the agent states. 
Unlike models that apply displacements directly in the global frame, our approach defines actions in the \textit{local coordinate frame} of the agent, which aligns with its heading direction.
At each timestep $t$, the action is represented as
\begin{equation}
a_t = (\Delta x_t, \Delta y_t, \Delta \psi_t),
\end{equation}
where $\Delta x_t$ and $\Delta y_t$ denote the forward and lateral displacements relative to the agent’s orientation, and $\Delta \psi_t$ is the change in yaw angle.
To apply the action in the global coordinate frame, we rotate the local displacement using the current yaw $\psi_t$ of the agent:
\begin{equation}
\begin{bmatrix}
\Delta x_t^{\text{global}} \\
\Delta y_t^{\text{global}}
\end{bmatrix}
=
R(\psi_t)
\begin{bmatrix}
\Delta x_t \\
\Delta y_t
\end{bmatrix},
\quad
\Delta \psi_t^{\text{global}} = \Delta \psi_t,
\end{equation}

where $R(\psi_t)$ is the 2D rotation matrix. The global displacements are then added to the agent’s current state $(x_t, y_t, \psi_t)$ to obtain the updated trajectory $(x_{t+1}, y_{t+1}, \psi_{t+1})$.
For inverse dynamics, the procedure is reversed: we compute the displacement between consecutive states in the global frame and project it back into the local frame using $R(-\psi_t)$.  
This formulation allows actions to be expressed relative to the agent’s forward-facing direction, making them more natural for imitation learning and policy optimization tasks.

\subsubsection{Bicycle Model}
\label{app:bicycle}
Following the kinematic bicycle model \citep{gulino2023waymax}, we define an agent's current state information as $s = (x, y, \theta, v_x, v_y)$, which includes the $x,y$ positions in the coordinate space, the yaw angle $\theta$, and the velocities in the $X$ and $Y$ directions. The action is defined as $(a, \kappa)$, where $a$ denotes the longitudinal acceleration and $\kappa$ the steering curvature. Let $v=\sqrt{v_x^2+v_y^2}$ denote the speed magnitude.

For the inverse kinematics, given the state information of two consecutive states $s=(x, y, \theta, v_x, v_y)$ and $s'=(x', y', \theta', v_x', v_y')$, we estimate the acceleration $a$ and steering curvature $\kappa$ as
\[
a = \frac{v' - v}{\Delta t}
\]
\[
\kappa = \frac{\theta' - \theta}{d},
\qquad
d = \sqrt{(x'-x)^2 + (y'-y)^2},
\]
where $v'=\sqrt{{v_x'}^2 + {v_y'}^2}$. This formulation provides a compact approximation of car-like motion for extracting action labels from logged trajectories.

\subsection{Trajectory Types Distributions}
\label{app:trajectory_types}
\input{Figures/tex/full_type_dist}

\textit{Straight} refers to trajectories in which changes in $\Delta y$ and $\Delta yaw$ remain below predefined thresholds. \textit{Turn} denotes the presence of a contiguous interval with notable variations in $\Delta y$ and $\Delta yaw$. \textit{Reverse} captures cases where $\Delta x$ is negative for at least half of the trajectory. \textit{Uncategorized} encompasses all remaining trajectories that do not satisfy the criteria above. The thresholds and criteria employed for this categorization are empirically determined using domain knowledge and a preliminary inspection of the dataset. As shown in Figure \ref{fig:type-dist}, the trajectory distribution is highly imbalanced, with nominal scenarios—particularly \textit{Uncategorized} and \textit{Straight}—accounting for more than 80\% of the data, whereas \textit{Reverse} cases constitute less than 1\%.

To label the trajectory type, we extract the action values ($\Delta x,  \Delta y, \Delta yaw$) of whole trajectories. The training dataset has 542K samples, and the validation dataset has 69 K samples. Empirically, we first filtered out trajectories whose maximum lateral offset, $\max(\Delta y)$, exceeded 0.5 and whose maximum yaw change, $\max(\Delta \text{yaw})$, exceeded 0.2. We then labeled the remaining trajectories as follows: a trajectory was labeled \textbf{Straight} if the peaks of both $\Delta y$ and $\Delta \text{yaw}$ were below 0.01; \textbf{Turn} if the fraction of timesteps with $\Delta y > 0.035$ and $\Delta \text{yaw} > 0.025$ was at least 15\% of the trajectory; \textbf{Reverse} if the proportion of timesteps with $\Delta x < -0.01$ was at least 50\%; and \textbf{Uncategorized} otherwise.
\newpage

%% file: Tables/2_obs_features.tex
\renewcommand{\arraystretch}{1.2}
\begin{table}[h]
\centering
\caption{Observation features specifications.}
\label{tab:obs_features}
\resizebox{\columnwidth}{!}{
\begin{tabular}{|l|l|l|l|}
\hline
\textbf{Type} & \textbf{Feature} & \textbf{Dimension} & \textbf{Description} \\
\hline
\multirow{4}{*}{ego state} 
 & Speed & $(1,)$ & Ego vehicle speed \\
 & Size & $(2,)$ & Vehicle length and width \\
 & Goal position & $(2,)$ & Relative goal coordinates (x,y) \\
 & Collision state & $(1,)$ & 1 if collided, 0 otherwise \\
\hline
\multirow{4}{*}{surrounding vehicle} 
 & Speed & $(\text{max num agents} - 1, 1)$ & Partner vehicle speed \\
 & Size & $(\text{max num agents} - 1, 2)$ & Length and width \\
 & Goal position & $(\text{max num agents} - 1, 2)$ & Relative goal coordinates (x,y) \\
 & Collision state & $(\text{max num agents} - 1, 1)$ & 1 if collided, 0 otherwise \\
\hline
\multirow{4}{*}{road graph} 
 & Segment position & $(\text{top k road points}, 2)$ & Road point coordinates \\
 & Segment size & $(\text{top k road points}, 3)$ & Length, width, and height \\
 & Segment orientation & $(\text{top k road points}, 1)$ & Orientation of the road segment \\
 & Segment type & $(\text{top k road points}, 8)$ & One-hot encoded road point type \\
\hline
\end{tabular}%
}
\end{table}

%% file: Figures/tex/full_type_dist.tex
\begin{wrapfigure}{r}{0.47\columnwidth}
    \vspace{-20pt}
    \centering  
    \includegraphics[width=\linewidth]{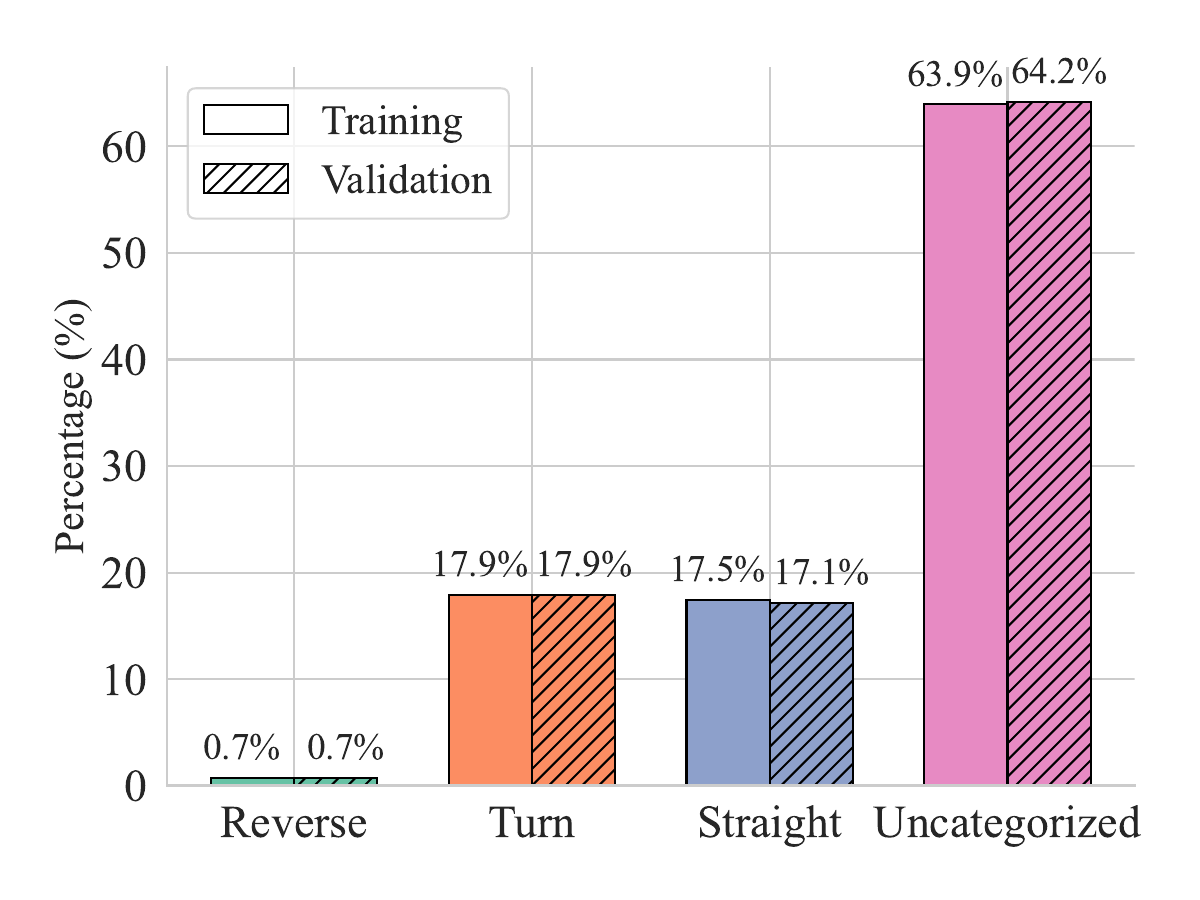}
    \vspace{-18pt}
    \caption{\textbf{Distributions of 4 trajectory types on training and validation set.}}
    \label{fig:type-dist}
\end{wrapfigure}

%% file: appendix/c_scaling.tex
\section{Additional Scaling Results}

\subsection{Training and Validation Scenes Results}
\input{Tables/4_driving_performance_metrics}
\label{app:train-validation}
As shown in Table \ref{tab:performance-metrics_bc}, the BC model improves consistently on almost all metrics up to 10,000 scenes, after which performance largely saturates. The generalization gap between the training and validation sets is also mostly resolved beyond 20,000 scenes. By contrast, for RL, Table \ref{tab:performance-metrics_rl} shows gradual improvement in all metrics except the goal progress ratio. This indicates a weaker scaling trend than in BC, likely because the RL objective is more directly tied to destination-reaching behavior.

\subsection{WOSAC metrics of Power-law Relationship}
\label{app:wosac}
\input{Figures/tex/wosac}
WOSAC metrics are widely used to evaluate human likeness in autonomous driving. Following this, we run a data-scaling study to test whether planning quality exhibits a power-law relationship with the amount of training data. As shown in Figure ~\ref{fig:wosac-full}, all five metrics are strongly correlated with the number of scenes, and are well-approximated by a power-law fit. The four WOSAC sub-metrics—Realism Meta, Kinematic, Interactive, and Map-based—improve steadily with data up to $\sim$ 10k scenes, after which gains largely saturate. Our Realism Meta score approaches 0.7, and minADE reaches $\sim$ 1.1, both within a reasonable range and close to the 2023 leaderboard. This saturation behavior mirrors our surrounding-vehicle linear probing results, which also plateau beyond $\sim$ 10K scenes. Finally, our random baseline matches that reported in PufferDrive \citep{pufferdrive2025github}, providing a reference point for interpreting the absolute scale of these scores.

\subsection{Additional results of Power-law Relationship}
\label{app:scale-law}
\input{Figures/tex/fule_power_law}
\input{Figures/tex/power_law_cases_bc}
In this section, we show the additional power-law relationship results for closed-loop metrics as shown in Figure \ref{fig:scaling-law-closed} and for open-loop metrics as shown in Figure \ref{fig:scaling-law-open}.
To show the performance by cases, we also conduct the power-law relationship by types as in Figure \ref{fig:power-law-types}. In most cases, the coefficient $r$ was high, indicating a strong relationship between data scale and performance. In the case of $Reverse$, the metrics have a high standard deviation because only rare cases occur. However, there are no collisions after 5,000 scenes.

\newpage
\clearpage

%% file: Tables/4_driving_performance_metrics.tex
\begin{table}[h]
\caption{Driving performance metrics across dataset scales of BC model.}
\label{tab:performance-metrics_bc}
\centering
\resizebox{\columnwidth}{!}{%
\begin{tabular}{c|l|cccc}
\hline
\makecell{\textbf{Num}\\ \textbf{Scenes}} & \textbf{Dataset} & \textbf{Goal Rate} & \textbf{Off-Road} & \textbf{Veh-Coll} & \makecell{\textbf{Goal Progress}\\ \textbf{Ratio}} \\
\hline
\multirow{2}{*}{\centering 100} & Training & 0.320 $\pm$ 0.062 & 0.306 $\pm$ 0.050 & 0.125 $\pm$ 0.031 & 0.606 $\pm$ 0.149 \\
 & Validation & 0.302 $\pm$ 0.034 & 0.352 $\pm$ 0.030 & 0.183 $\pm$ 0.011 & 0.561 $\pm$ 0.115 \\
\cline{1-6}
\multirow{2}{*}{\centering 500} & Training & 0.568 $\pm$ 0.133 & 0.127 $\pm$ 0.052 & 0.092 $\pm$ 0.012 & 0.785 $\pm$ 0.092 \\
 & Validation & 0.521 $\pm$ 0.111 & 0.156 $\pm$ 0.033 & 0.111 $\pm$ 0.018 & 0.745 $\pm$ 0.083 \\
\cline{1-6}
\multirow{2}{*}{\centering 1000} & Training & 0.553 $\pm$ 0.038 & 0.074 $\pm$ 0.019 & 0.083 $\pm$ 0.025 & 0.780 $\pm$ 0.063 \\
 & Validation & 0.543 $\pm$ 0.017 & 0.114 $\pm$ 0.011 & 0.107 $\pm$ 0.024 & 0.748 $\pm$ 0.072 \\
\cline{1-6}
\multirow{2}{*}{\centering 5000} & Training & 0.572 $\pm$ 0.097 & 0.074 $\pm$ 0.014 & 0.070 $\pm$ 0.011 & 0.797 $\pm$ 0.019 \\
 & Validation & 0.573 $\pm$ 0.093 & 0.086 $\pm$ 0.013 & 0.073 $\pm$ 0.007 & 0.795 $\pm$ 0.015 \\
\cline{1-6}
\multirow{2}{*}{\centering 10000} & Training & 0.615 $\pm$ 0.167 & 0.058 $\pm$ 0.008 & 0.076 $\pm$ 0.020 & 0.808 $\pm$ 0.043 \\
 & Validation & 0.615 $\pm$ 0.166 & 0.067 $\pm$ 0.008 & 0.073 $\pm$ 0.019 & 0.809 $\pm$ 0.041 \\
\cline{1-6}
\multirow{2}{*}{\centering 20000} & Training & 0.710 $\pm$ 0.020 & 0.075 $\pm$ 0.031 & 0.062 $\pm$ 0.006 & 0.854 $\pm$ 0.039 \\
 & Validation & 0.709 $\pm$ 0.018 & 0.077 $\pm$ 0.030 & 0.060 $\pm$ 0.002 & 0.857 $\pm$ 0.035 \\
\cline{1-6}
\multirow{2}{*}{\centering 40000} & Training & 0.567 $\pm$ 0.163 & 0.067 $\pm$ 0.006 & 0.073 $\pm$ 0.034 & 0.836 $\pm$ 0.076 \\
 & Validation & 0.566 $\pm$ 0.159 & 0.069 $\pm$ 0.010 & 0.071 $\pm$ 0.036 & 0.839 $\pm$ 0.073 \\
 \cline{1-6}
\multirow{2}{*}{\makecell{80000\\(Full)}} & Training & 0.583 $\pm$ 0.033 & 0.066 $\pm$ 0.004 & 0.076 $\pm$ 0.014 & 0.808 $\pm$ 0.074 \\
 & Validation & 0.587 $\pm$ 0.034 & 0.065 $\pm$ 0.003 & 0.073 $\pm$ 0.015 & 0.813 $\pm$ 0.072 \\
\hline
\end{tabular}%
}
\end{table}

\begin{table}[h]
\caption{Driving performance metrics across dataset scales of the RL model.}
\label{tab:performance-metrics_rl}
\centering
\resizebox{\columnwidth}{!}{%
\begin{tabular}{c|l|cccc}
\hline
\makecell{\textbf{Num}\\ \textbf{Scenes}} & \textbf{Dataset} & \textbf{Goal Rate} & \textbf{Off-Road} & \textbf{Veh-Coll} & \makecell{\textbf{Goal Progress}\\ \textbf{Ratio}} \\
\hline
100 & Validation & 0.491 $\pm$ 0.123 & 0.088 $\pm$ 0.034 & 0.091 $\pm$ 0.020 & 0.519 $\pm$ 0.119 \\
\cline{1-6}
500 & Validation & 0.743 $\pm$ 0.058 & 0.081 $\pm$ 0.010 & 0.069 $\pm$ 0.012 & 0.757 $\pm$ 0.067 \\
\cline{1-6}
1000 & Validation & 0.875 $\pm$ 0.038 & 0.077 $\pm$ 0.017 & 0.060 $\pm$ 0.008 & 0.885 $\pm$ 0.036 \\
\cline{1-6}
5000 & Validation & 0.861 $\pm$ 0.101 & 0.068 $\pm$ 0.011 & 0.040 $\pm$ 0.003 & 0.882 $\pm$ 0.072 \\
\cline{1-6}
10000 & Validation & 0.989 $\pm$ 0.005 & 0.034 $\pm$ 0.011 & 0.023 $\pm$ 0.005 & 0.986 $\pm$ 0.005 \\
\hline
\end{tabular}%
}
\end{table}

%% file: Figures/tex/wosac.tex
\begin{figure*}[h]
    \centering
    \includegraphics[width=\linewidth]{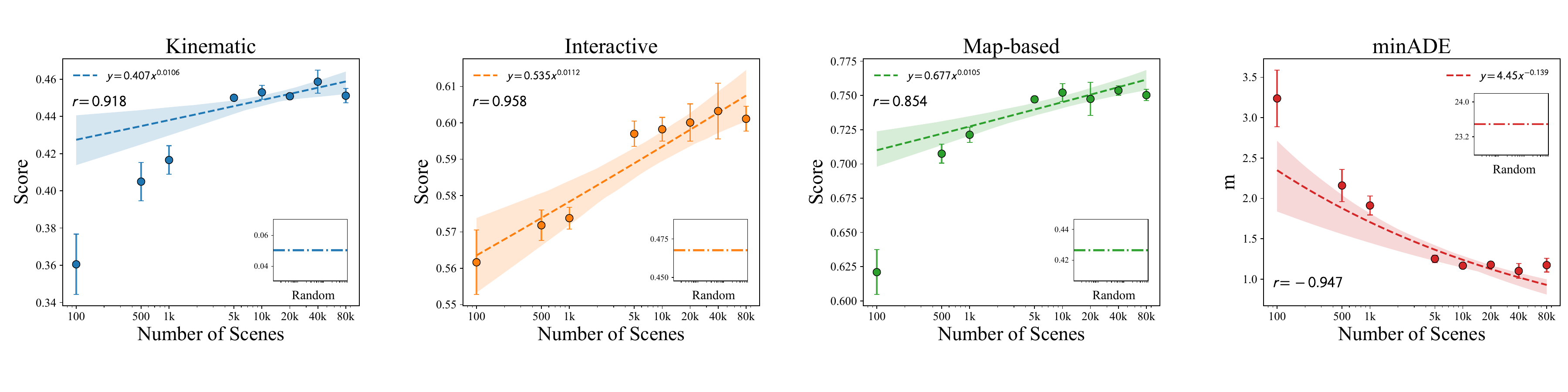}
    \caption{\textbf{Power law relationships for WOSAC metrics}: We evaluate the realism meta score, kinematic score, interactive score, map-based score, and minADE for 1,000 scenes in the validation set.}
    \label{fig:wosac-full}
\end{figure*}

%% file: Figures/tex/fule_power_law.tex
\begin{figure}[h]
    \centering
    \begin{subfigure}[h]{0.49\columnwidth}
        \centering
        \includegraphics[width=\linewidth]{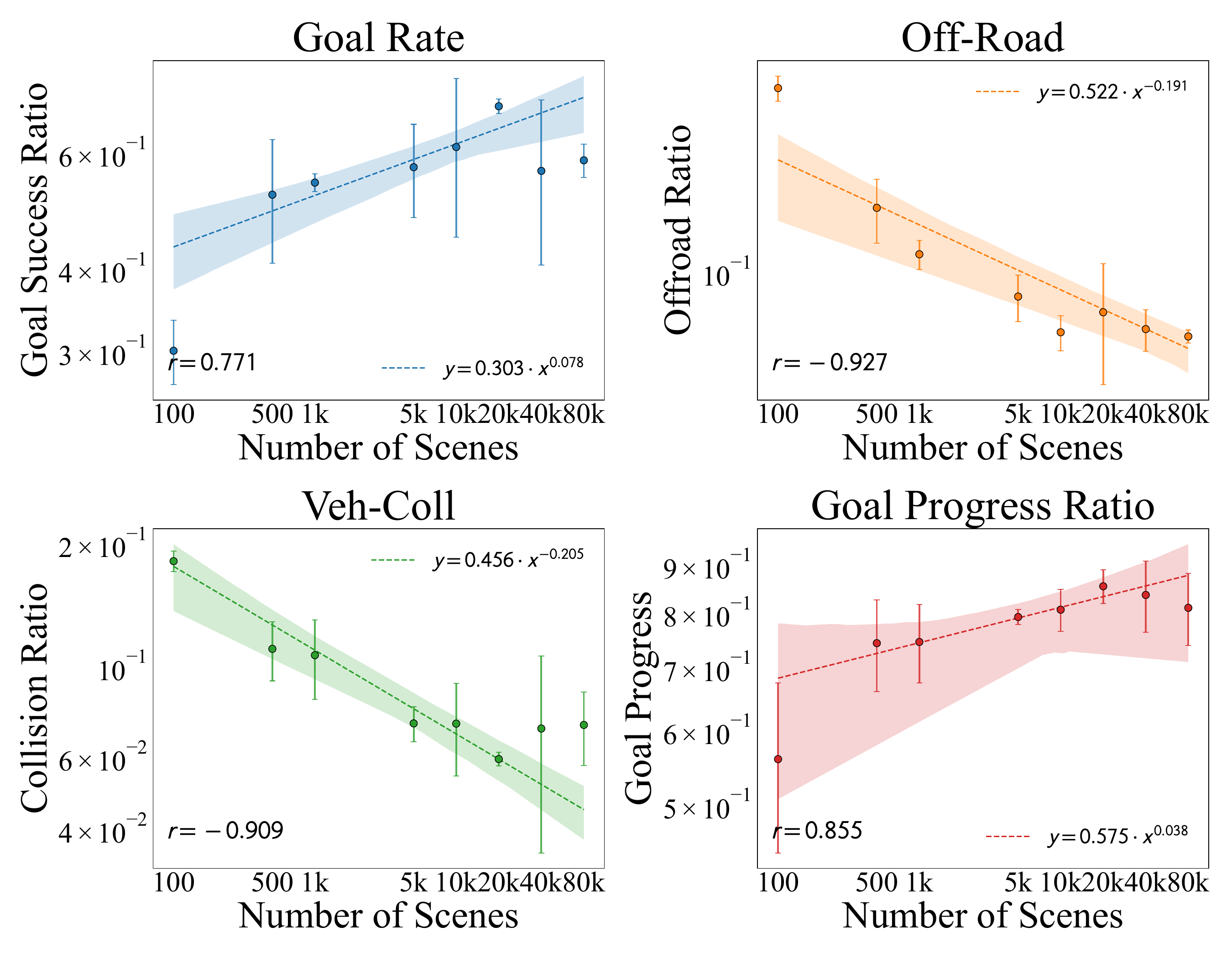}
        \caption{\textbf{Closed-loop scaling.} Goal/collision vs data; dashed: power-law fit; $r$: correlation.}
        \label{fig:scaling-law-closed}
    \end{subfigure}\hfill
    \begin{subfigure}[h]{0.49\columnwidth}
        \centering
        \includegraphics[width=\linewidth]{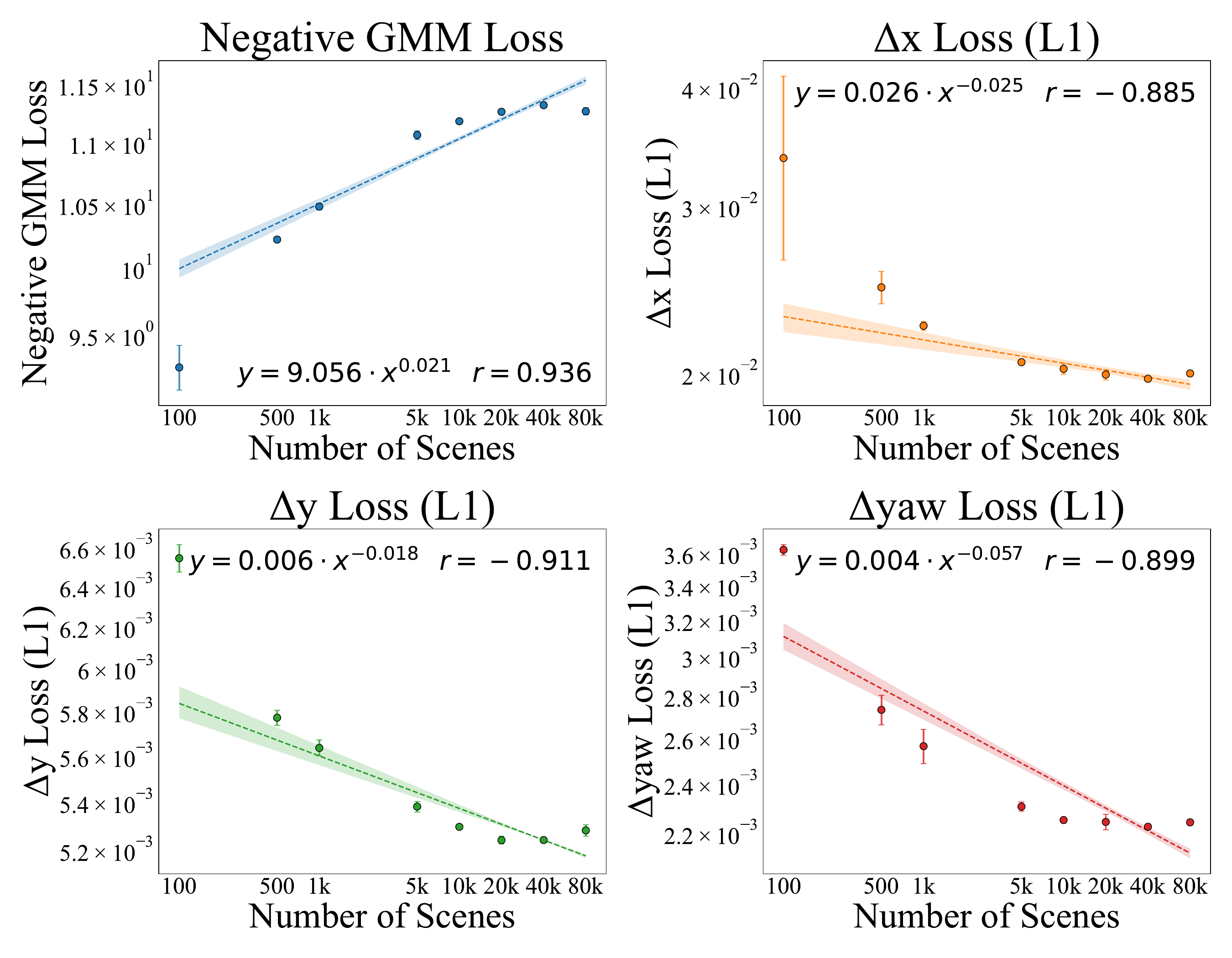}
        \caption{\textbf{Open-loop scaling.} GMM and action L1 losses vs data.}
        \label{fig:scaling-law-open}
    \end{subfigure}
\end{figure}

%% file: Figures/tex/power_law_cases_bc.tex
\begin{figure*}[th]
    \centering
    \includegraphics[width=\linewidth]{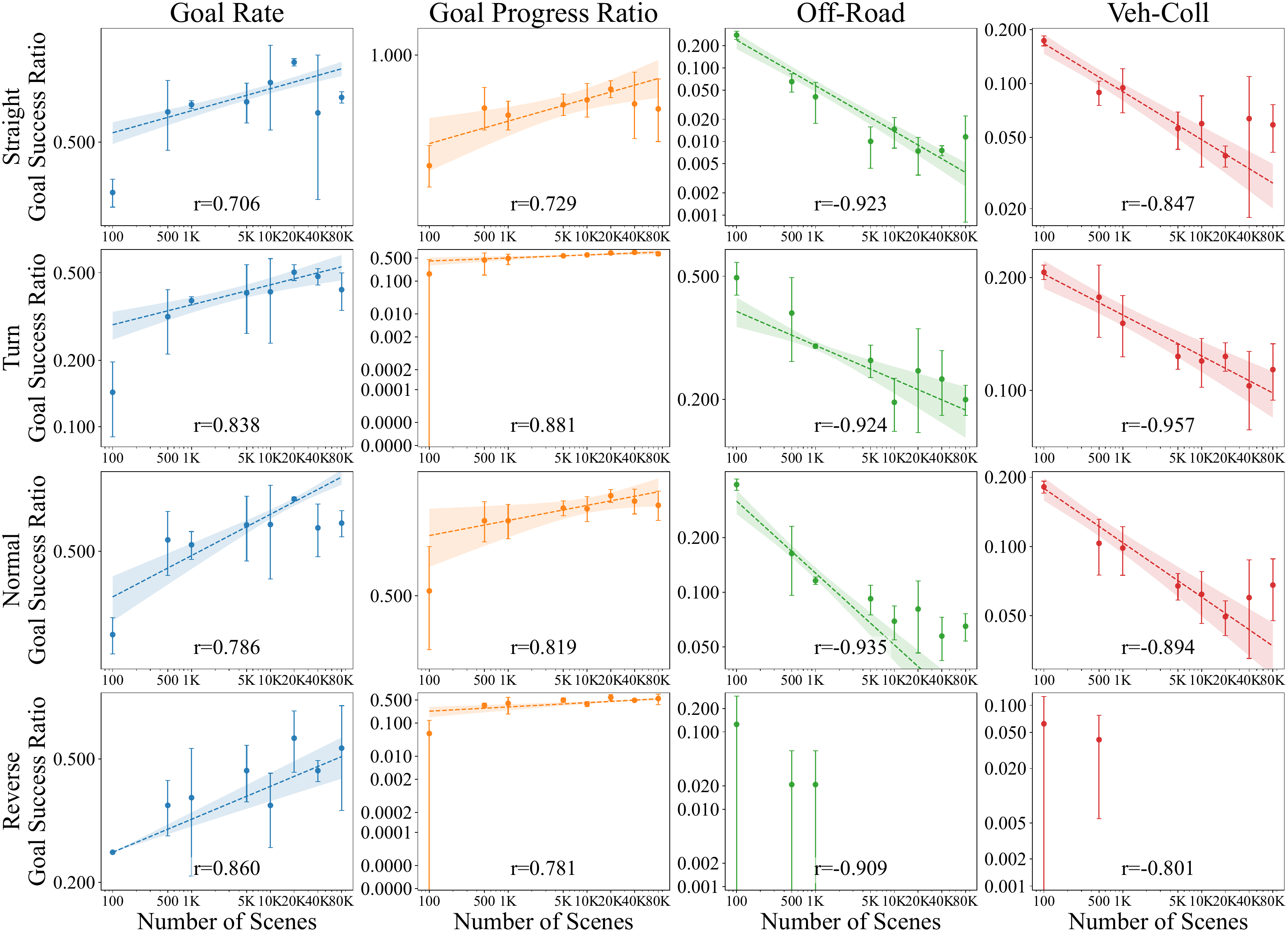}
    \caption{\textbf{Power law relationships for simulation results by cases (BC)}: Closed-loop simulation results for BC with data scaling. Missing points correspond to zero-valued metrics.}
    \label{fig:power-law-types}
\end{figure*}

%% file: appendix/d_surrounding.tex
\section{Additional Surrounding-vehicle probing results}
\label{app:other-probing}
\subsection{Linear Probing Setting}
\label{app:lp-other-setting}
\input{Tables/9_lp_other_setting}
Across all experiments with varying dataset sizes, we train with a learning rate of $0.0015$ and a batch size of 256. The settings across dataset sizes are shown in Table \ref{tab:probing-setting-surrounding}.a
For SMART (IL), we trained the linear probe for 15,000 gradient steps with the learning rate of $0.001$ and a batch size of 16. In the case of the raw-input model, we use agent-centric features: position, heading, velocity, and type with 11 historical steps.

\subsection{Detailed Results of Linear Probing}
\label{app:detail-other-probing}
\input{Figures/tex/other_probe_full}
In this section, we present the full linear probing results for BC and RL, shown in Figure \ref{fig:other-probing-bc} and Figure \ref{fig:other-probing-rl}. The BC model does learn about surrounding vehicles as the training scale increases, but its absolute predictive accuracy remains limited. Although both early and late LP consistently outperform the raw-input probe, the overall F1 score remains below 10\%, indicating weak predictive power for surrounding vehicles. This weakness is also reflected in type accuracy: BC performs better on relatively easy cases such as \textit{Straight} and \textit{Uncategorized}, but degrades substantially on more complex cases such as \textit{Turn} and \textit{Reverse}, suggesting that it prioritizes easier agents rather than the safety-critical ones. Interestingly, as the dataset grows, BC probing gains concentrate more on the near horizon, whereas the raw-input probe continues to improve at farther future steps. This suggests that the BC model increasingly focuses on imminent, collision-prone intervals while paying less attention to longer-range futures. The widening gap between early and late LP at larger scales further indicates that deeper representations increasingly encode near-future behavior.

By contrast, the RL model achieves consistently higher F1 scores than BC across all future horizons, suggesting that it learns stronger representations of surrounding-vehicle behavior. The fact that the raw-input probe is also stronger than in BC further suggests that self-play trajectories themselves are more predictable, indicating that RL agents may interact in a more mutually predictable manner. However, this advantage is concentrated on the near future. As the dataset grows, RL largely maintains its performance at the 10-step horizon, while showing weaker gains or slight degradation at longer horizons. This implies that RL increasingly prioritizes short-horizon behavior and discards information about the distant future. One possible explanation is that, unlike BC, which uses stacked observations over multiple past steps, RL operates on single-step inputs and therefore has less access to temporal context for longer-horizon prediction. Overall, these results suggest that RL learns stronger, more short-horizon-oriented representations of surrounding vehicles than BC does.

\input{Tables/5_f1_macro_other}
\input{Tables/6_info_loss_other}

Additionally, we introduce the additional results of linear probing as in the Table \ref{tab:f1_diff_other_column} and Table \ref{tab:acc_diff_types_other}. Note that the RL model does not have a late LP, since it fuses at an earlier layer. From the Table \ref{tab:f1_diff_other_column}, we report the F1 Macro score across all future timesteps. In Table \ref{tab:acc_diff_types_other}, we report the full results of type accuracy.

As shown in Table \ref{tab:f1_diff_other_column}, the BC model shows clear saturation after 10,000 scenes in both \textit{Info Loss} and \textit{LP - Raw}. Both values also decrease as the future step increases, indicating that the gain from surrounding-vehicle representations is concentrated on the near horizon. The positive \textit{Info Loss} further suggests that earlier layers retain more surrounding-vehicle information, while later layers gradually discard it, likely in favor of other signals such as ego planning.
By contrast, the RL model exhibits a much larger \textit{LP - Raw} gap even at 100 scenes, indicating that useful surrounding-vehicle representations emerge much earlier than in BC. This may help explain why RL shows relatively stable collision behavior. However, this advantage is concentrated at short horizons, while longer-horizon gains shrink substantially as scale increases.

As shown in Table \ref{tab:acc_diff_types_other}, the BC model exhibits increasingly larger type-wise probing gains with scale, especially for \textit{Normal} and \textit{Straight} cases, while \textit{Turn} remains consistently more difficult and \textit{Reverse} stays weak and unstable. The positive \textit{Info Loss} across most types further suggests that earlier layers retain more surrounding-vehicle information, which is gradually discarded in later layers. By contrast, RL shows relatively strong \textit{LP - Raw} gains even at small scales, indicating earlier emergence of useful surrounding-vehicle representations. However, unlike in BC, these gains do not increase monotonically with scale; instead, they weaken at 10K scenes for several types. Overall, BC improves more steadily with data, whereas RL learns useful type-specific representations earlier but does not sustain the same scaling trend.

\subsection{Type Accuracy of Future Steps}
\label{app:type-other-probing}

\begin{figure*}[th]
    \centering
    \includegraphics[width=0.9\linewidth]{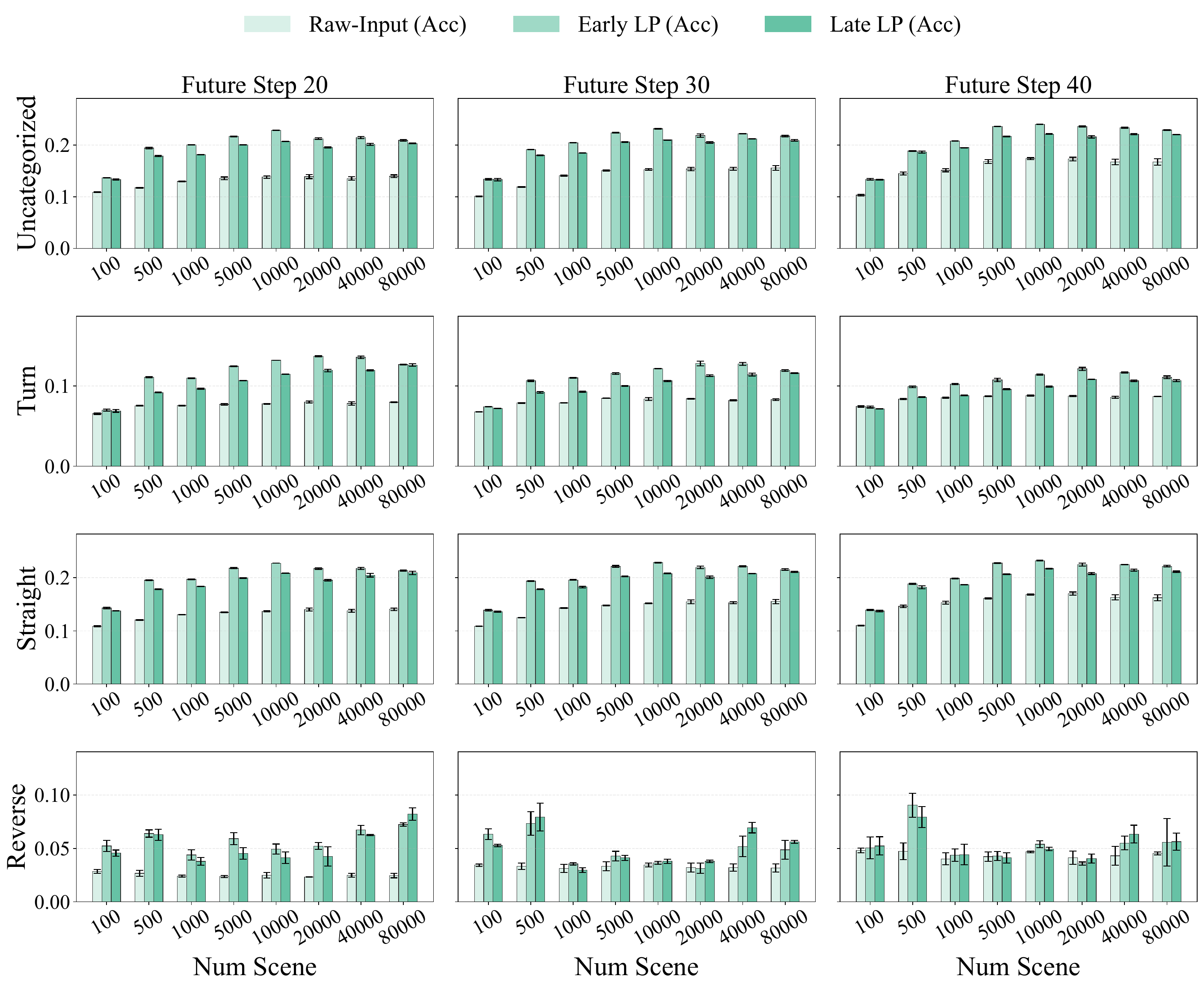}
    \caption{\textbf{Type accuracy of other future steps (20, 30, 40) (Other) (BC)}: The row is type ($Normal$, $Turn$, $Straight$, and $Reverse$) and the column is future step (20, 30, 40).}
    \label{fig:other-types}
\end{figure*}

\begin{figure*}[th]
    \centering
    \includegraphics[width=0.9\linewidth]{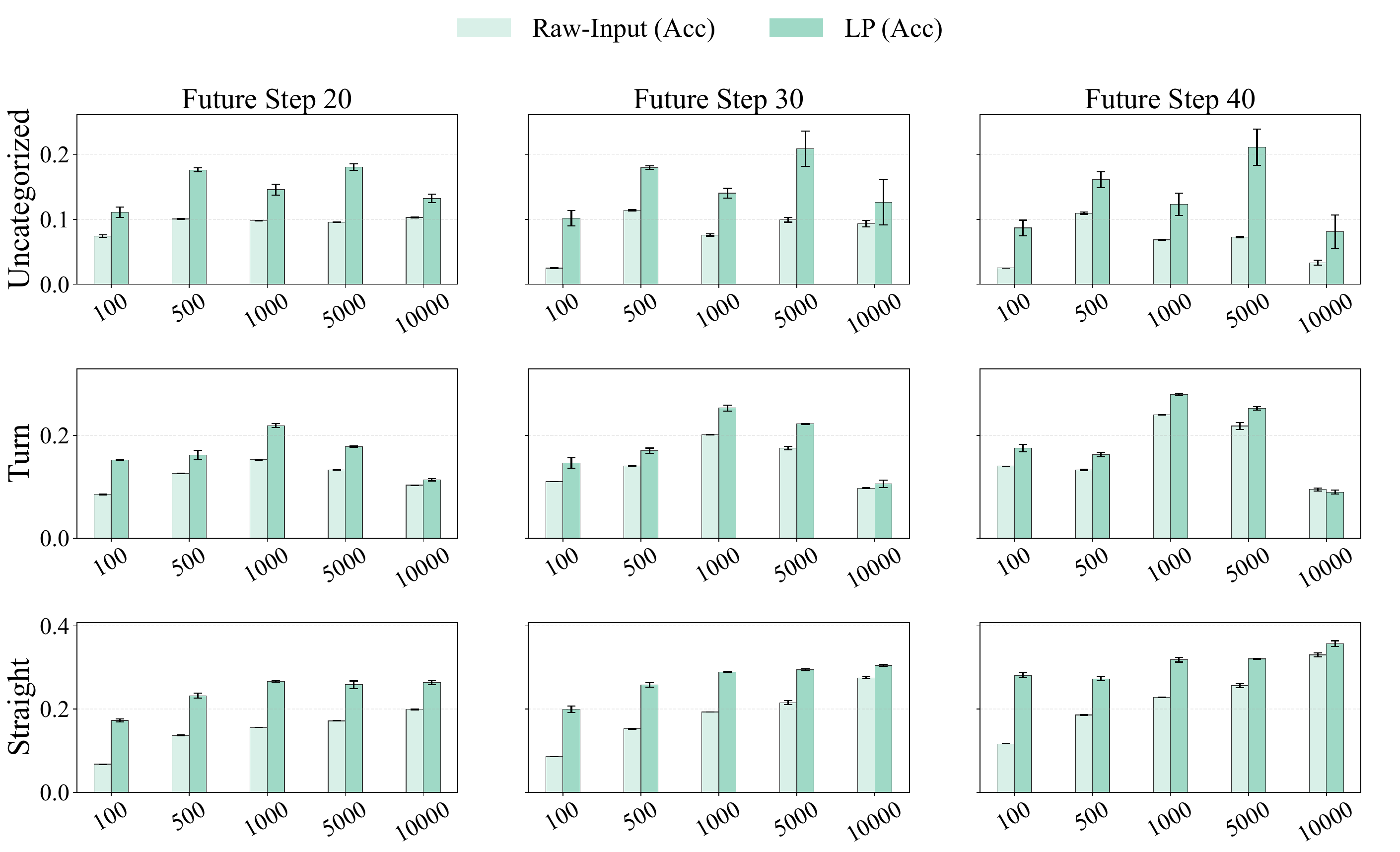}
    \caption{\textbf{Type accuracy of other future steps (20, 30, 40) (Other) (RL)}: The row is type ($Normal$, $Turn$, $Straight$, and $Reverse$) and the column is future step (20, 30, 40).}
    \label{fig:other-types-rl}
\end{figure*}

The figure \ref{fig:other-types} shows the corresponding results for surrounding-vehicle probing. Here, accuracies are overall lower than in the ego case, and the gap between early and late LP is smaller and less stable, especially for $Turn$ and $Reverse$. However, there is no tendency for type accuracy in RL as in \ref{fig:other-types-rl}.

\newpage
\clearpage

%% file: Tables/9_lp_other_setting.tex
\begin{table}[!htbp]
\caption{Linear Probing Setting Details for Surrounding-Vehicle Prediction.}
\label{tab:probing-setting-surrounding}
\centering
\resizebox{0.75\columnwidth}{!}{%
\begin{tabular}{c|c|c|c|c|c}
\hline
\textbf{\#Scenes} & \makecell{\textbf{Gradient}\\\textbf{(Step)}} & \makecell{\textbf{\#Samples}\\\textbf{(fs@10)}} & \makecell{\textbf{\#Samples}\\\textbf{(fs@20)}} & \makecell{\textbf{\#Samples}\\\textbf{(fs@30)}} & \makecell{\textbf{\#Samples}\\\textbf{(fs@40)}}\\
\hline
100    & 5K    & 235K   & 184K   & 140K   & 104K \\
500    & 12K   & 983K   & 768K   & 588K   & 437K \\
1000   & 20K   & 1.8M   & 1.4M   & 1.1M   & 824K \\
5000   & 50K   & 8.6M   & 6.7M   & 5.1M   & 3.8M \\
10,000 & 75K   & 17.6M  & 13.7M  & 10.5M  & 7.8M \\
20,000 & 100K  & 35.2M  & 27.4M  & 21M    & 15.7M \\
40,000 & 125K  & 70M    & 54.6M  & 41.9M  & 31.2M \\
80,000 & 150K  & 140.4M & 110M   & 84.1M  & 62.7M \\
\hline
\end{tabular}%
}
\end{table}

%% file: Figures/tex/other_probe_full.tex
\begin{figure*}[h]
    \vspace{-3mm}
    \centering
    \includegraphics[width=\linewidth]{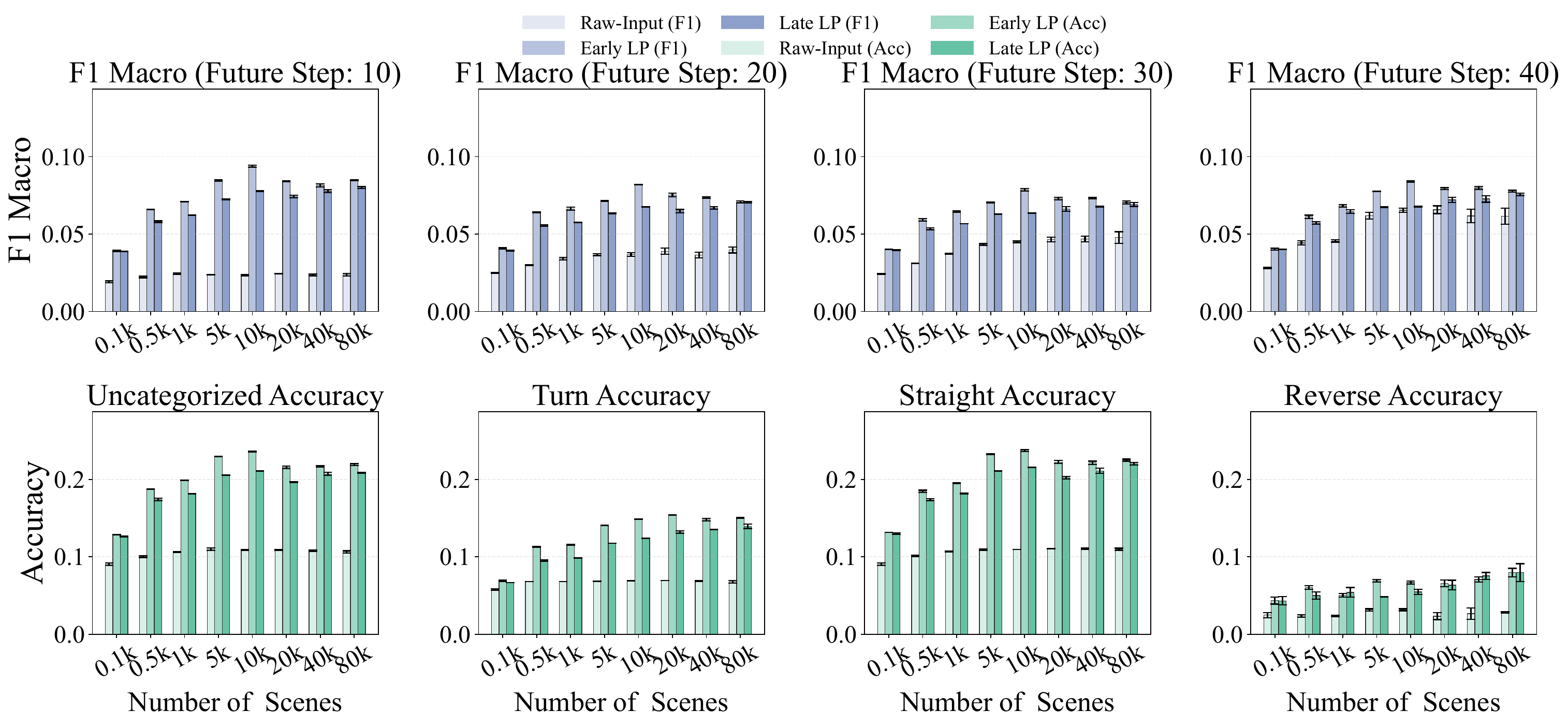}
    \caption{\textbf{Full performance metrics with surrounding-linear probing (BC)}: Light colors are raw-input probing and darker colors are BC probing(earlier layer and late layer) with error bars of standard deviation across different seeds. For convenience, we refer to linear probing in the earlier layer of the BC model as \textit{early LP} and in the later layer as \textit{late LP}. The top row is the F1 score across future step $\{10, 20, 30, 40\}$ and the bottom row is accuracy with labeled cases (Future step = 10).}
    \label{fig:other-probing-bc}
\end{figure*}

\begin{figure*}[h]
    \vspace{-3mm}
    \centering
    \includegraphics[width=\linewidth]{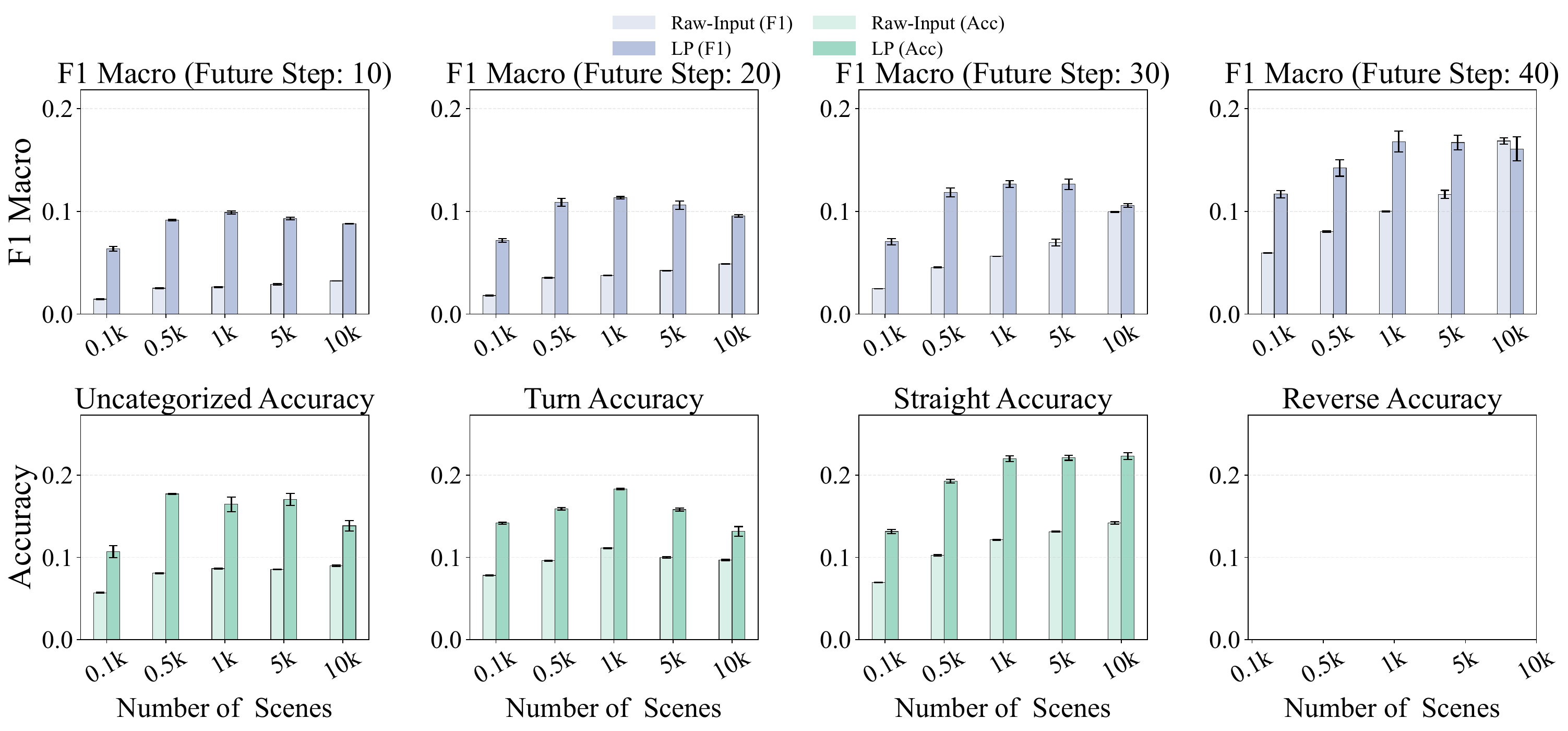}
    \caption{\textbf{Full performance metrics with surrounding-linear probing (RL)}: Light colors are raw-input probing and darker colors are RL probing with error bars of standard deviation across different seeds. For convenience, we refer to linear probing as \textit{LP}. The top row is the F1 score across future steps $\{10, 20, 30, 40\}$, and the bottom row is accuracy with labeled cases (Future step = 10). Note that since there are no success cases of $Reverse$, so no probing results in $Reverse$.}
    \label{fig:other-probing-rl}
\end{figure*}

%% file: Tables/5_f1_macro_other.tex
\begin{table}[!htbp]
\centering
\caption{F1 Macro differences across future steps. (a) BC, (b) RL. $Info\ Loss$ is (Early LP $-$ Late LP), and $LP - Raw$ is the difference between probing and the raw-input model. The RL table only has $LP - Raw$.}
\label{tab:f1_diff_other_column}

\begin{subtable}{\columnwidth}
\centering
\caption{BC}
\label{tab:f1_diff_other_bc}
\resizebox{\linewidth}{!}{
\begin{tabular}{lrrrrrrrr}
\toprule
{} & \multicolumn{2}{c}{FS=10} & \multicolumn{2}{c}{FS=20} & \multicolumn{2}{c}{FS=30} & \multicolumn{2}{c}{FS=40} \\
{} & Info Loss. & LP - Raw & Info Loss. & LP - Raw & Info Loss. & LP - Raw & Info Loss. & LP - Raw \\
\midrule
100 & 0.0005 & 0.0199 & 0.0013 & 0.0158 & 0.0007 & 0.0160 & 0.0001 & 0.0124 \\
500 & 0.0080 & 0.0435 & 0.0086 & 0.0343 & 0.0061 & 0.0280 & 0.0039 & 0.0168 \\
1000 & 0.0088 & 0.0467 & 0.0086 & 0.0323 & 0.0078 & 0.0272 & 0.0037 & 0.0226 \\
5000 & 0.0121 & 0.0608 & 0.0080 & 0.0349 & 0.0077 & 0.0271 & 0.0099 & 0.0156 \\
10000 & 0.0161 & 0.0704 & 0.0144 & 0.0451 & 0.0151 & 0.0337 & 0.0162 & 0.0185 \\
20000 & 0.0099 & 0.0597 & 0.0103 & 0.0363 & 0.0066 & 0.0264 & 0.0074 & 0.0138 \\
40000 & 0.0038 & 0.0579 & 0.0067 & 0.0371 & 0.0057 & 0.0265 & 0.0071 & 0.0181 \\
80000 & 0.0046 & 0.0610 & 0.0002 & 0.0310 & 0.0012 & 0.0227 & 0.0023 & 0.0163 \\
\bottomrule
\end{tabular}
}
\end{subtable}

\begin{subtable}{\columnwidth}
\centering
\caption{RL}
\label{tab:f1_diff_other_rl}
\resizebox{0.55\linewidth}{!}{
\begin{tabular}{lcccc}
\toprule
\#Scenes & FS=10 & FS=20 & FS=30 & FS=40 \\
\midrule
100 & 0.0490 & 0.0537 & 0.0459 & 0.0573 \\
500 & 0.0665 & 0.0735 & 0.0731 & 0.0621 \\
1000 & 0.0726 & 0.0757 & 0.0703 & 0.0679 \\
5000 & 0.0641 & 0.0637 & 0.0567 & 0.0505 \\
10000 & 0.0558 & 0.0468 & 0.0063 & -0.0077 \\
\bottomrule
\end{tabular}
}
\end{subtable}
\end{table}

%% file: Tables/6_info_loss_other.tex
\begin{table}[!htbp]
\centering
\caption{Accuracy differences by action types. (a) BC at future step 10, (b) RL at future step 10. $Info\ Loss$ is (Early LP $-$ Late LP), and $LP - Raw$ is the difference between probing and the raw-input model.}
\label{tab:acc_diff_types_other}

\begin{subtable}{\columnwidth}
\centering
\caption{BC (Future step = 10)}
\label{tab:acc_diff_types_bc_other}
\resizebox{\linewidth}{!}{
\begin{tabular}{lrrrrrrrr}
\toprule
{} & \multicolumn{2}{c}{Normal} & \multicolumn{2}{c}{Turn} & \multicolumn{2}{c}{Straight} & \multicolumn{2}{c}{Reverse} \\
{} & Info Loss. & LP - Raw & Info Loss. & LP - Raw & Info Loss. & LP - Raw & Info Loss. & LP - Raw \\
\midrule
100 & 0.0021 & 0.0380 & 0.0025 & 0.0116 & 0.0020 & 0.0411 & 0.0004 & 0.0188 \\
500 & 0.0132 & 0.0872 & 0.0176 & 0.0451 & 0.0112 & 0.0838 & 0.0104 & 0.0368 \\
1000 & 0.0173 & 0.0925 & 0.0169 & 0.0474 & 0.0132 & 0.0882 & -0.0039 & 0.0306 \\
5000 & 0.0242 & 0.1198 & 0.0231 & 0.0722 & 0.0217 & 0.1233 & 0.0207 & 0.0373 \\
10000 & 0.0247 & 0.1271 & 0.0246 & 0.0798 & 0.0218 & 0.1279 & 0.0121 & 0.0352 \\
20000 & 0.0194 & 0.1070 & 0.0218 & 0.0845 & 0.0203 & 0.1120 & 0.0022 & 0.0424 \\
40000 & 0.0100 & 0.1090 & 0.0128 & 0.0793 & 0.0104 & 0.1111 & -0.0047 & 0.0488 \\
80000 & 0.0109 & 0.1127 & 0.0110 & 0.0826 & 0.0044 & 0.1151 & 0.0002 & 0.0516 \\
\bottomrule
\end{tabular}
}
\end{subtable}

\begin{subtable}{\columnwidth}
\centering
\caption{RL (Future step = 10)}
\label{tab:acc_diff_types_ego_other}
\resizebox{0.55\linewidth}{!}{
\begin{tabular}{lrrrr}
\toprule
{} & Normal & Turn & Straight & Reverse \\
{} & LP - Raw & LP - Raw & LP - Raw & LP - Raw \\
\midrule
100 & 0.0501 & 0.0638 & 0.0619 & -- \\
500 & 0.0965 & 0.0633 & 0.0905 & -- \\
1000 & 0.0783 & 0.0720 & 0.0988 & -- \\
5000 & 0.0850 & 0.0581 & 0.0900 & -- \\
10000 & 0.0484 & 0.0348 & 0.0815 & -- \\
\bottomrule
\end{tabular}
}
\end{subtable}
\end{table}

%% file: appendix/e_ego.tex
\section{Additional Ego probing results}
\label{app:ego-probing}
\subsection{Linear Probing Setting with Data Scaling}
\label{app:lp-ego-setting}
\input{Tables/9_lp_ego_setting}
Across all experiments with varying dataset sizes, we train with a learning rate of $0.0015$ and a batch size of 256. The different settings across different dataset sizes are in Table \ref{tab:probing-setting-ego}.

\subsection{Detailed Results of Linear Probing}
\label{app:detail-ego-probing}
\begin{figure*}[h]
    \centering
    \includegraphics[width=0.9\linewidth]{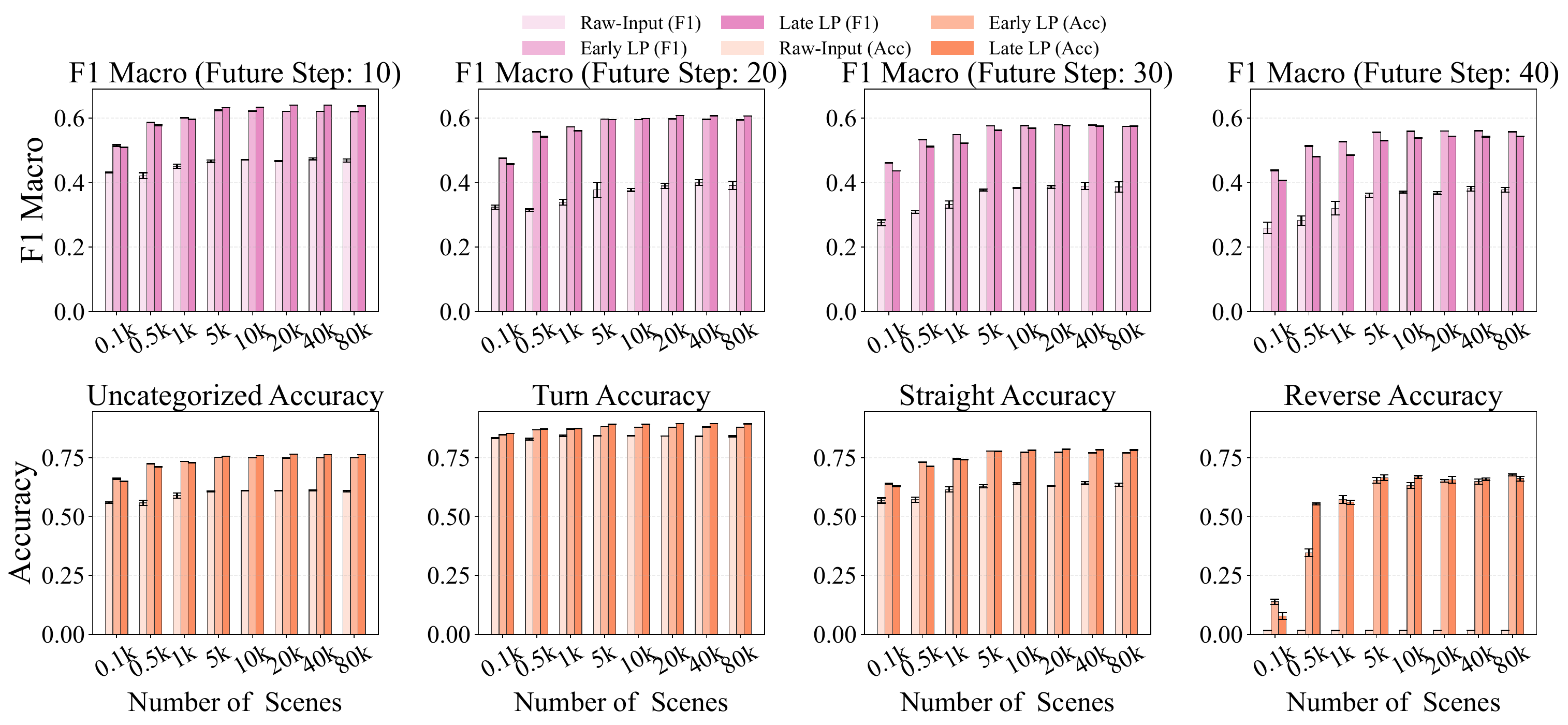}
    \caption{\textbf{Performance metrics with ego AV linear probing (BC)}: Light colors are raw-input probing and darker colors are BC probing (earlier layer and late layer) with error bars of standard deviation across different seeds. The top row is the F1 score across future steps $\{10, 20, 30, 40\}$ and the bottom row is accuracy with labeled cases (Future step = 10).}
    \label{fig:ego-probing-bc}
\end{figure*}

\begin{figure*}[h]
    \centering
    \includegraphics[width=0.9\linewidth]{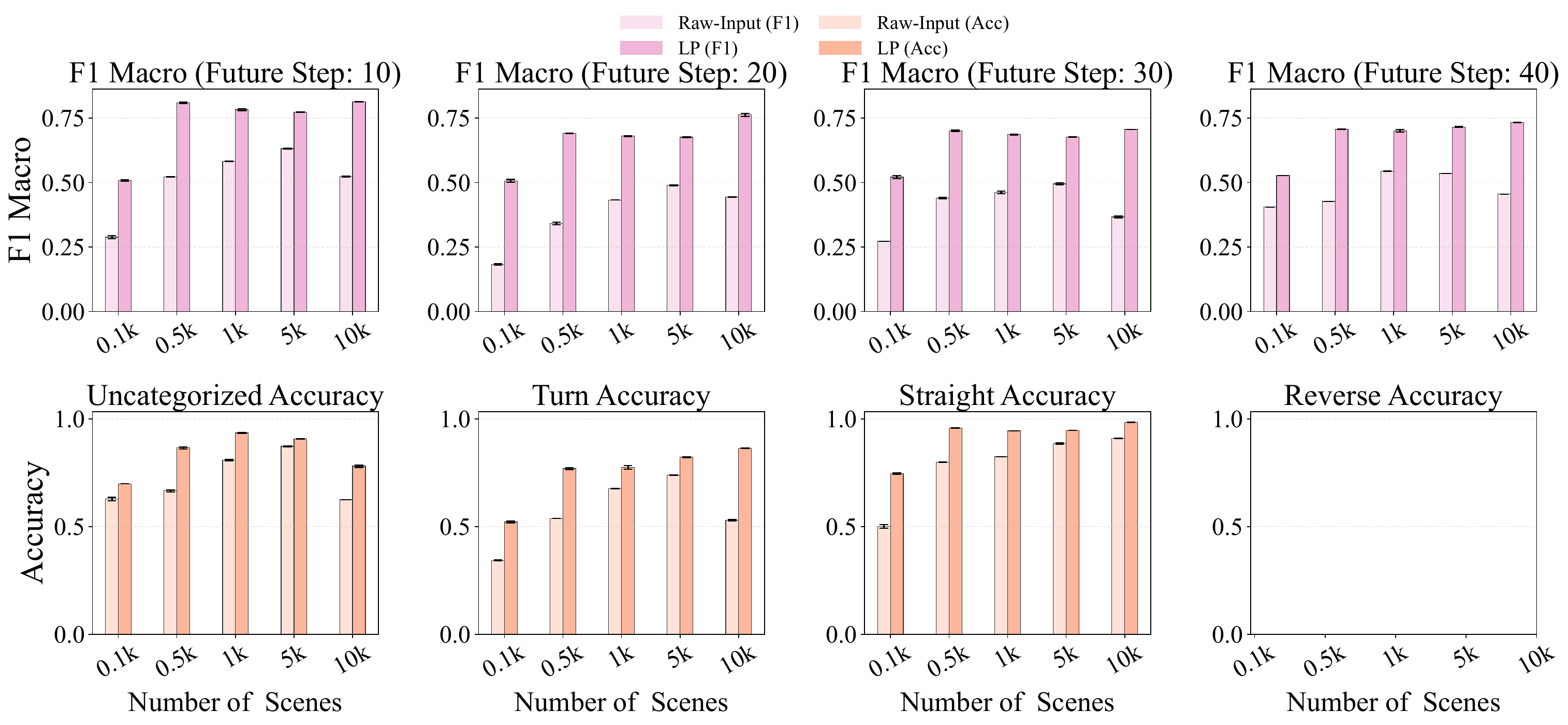}
    \caption{\textbf{Performance metrics with ego AV linear probing (RL)}: Light colors are raw-input probing and darker colors are RL probing (earlier layer and late layer) with error bars of standard deviation across different seeds. The top row is the F1 score across future steps $\{10, 20, 30, 40\}$ and the bottom row is accuracy with labeled cases (Future step = 10).}
    \label{fig:ego-probing-rl}
\end{figure*}

In this section, we report the full results of ego AV linear probing as in Figure \ref{fig:ego-probing-bc} and \ref{fig:ego-probing-rl}. The overall F1 Macro score exceeds 50\%, indicating that the BC model successfully learns not only about instant actions but also about long-horizon planning. As the dataset size increases, later-layer probes outperform early LP probes, suggesting that the BC model increasingly integrates map and other vehicle context when selecting actions. This trend contrasts with the linear probing of surrounding vehicles, where early LP probes surpass late LP probes (early LP $>$ late LP). Interestingly, the BC model forms coherent representations for reverse cases, even though such behavior is not reliably learned from raw inputs.

\input{Tables/5_f1_macro_ego}
\input{Tables/6_info_loss_ego}

\subsection{Type Accuracy of Future Steps}
\label{app:type-ego-probing}
\begin{figure*}[h]
    \centering
    \includegraphics[width=0.9\linewidth]{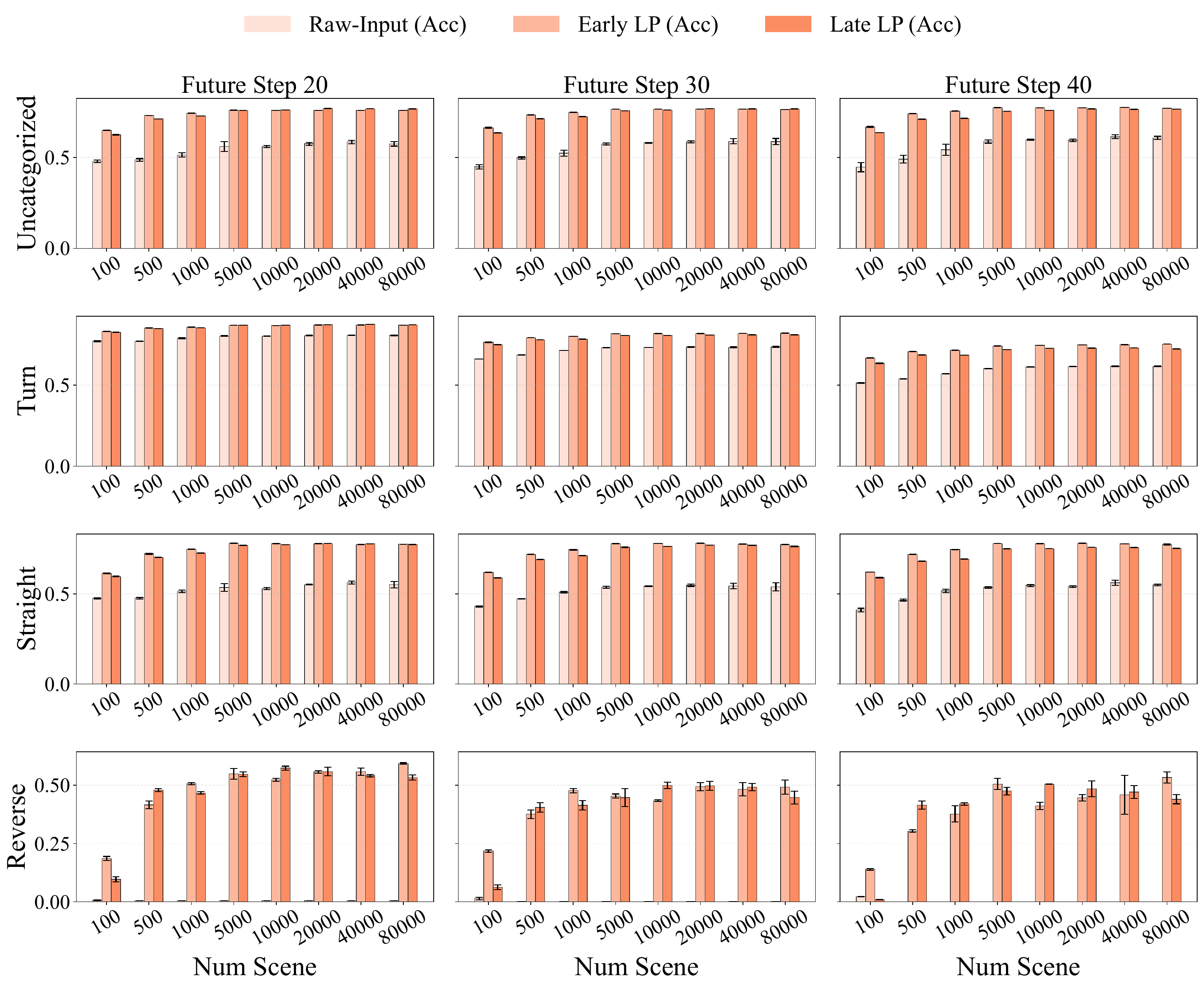}
    \caption{\textbf{Type accuracy of other future steps (20, 30, 40 (Ego) (BC)}: The row is type ($Normal$, $Turn$, $Straight$, and $Reverse$) and the column is future step (20, 30, 40).}
    \label{fig:ego-types}
\end{figure*}

The figure \ref{fig:ego-types} summarizes the accuracy of ego linear probing across trajectory types ($Uncategorized$, $Turn$, $Straight$, $Reverse$), future steps (20, 30, 40), and dataset scales. Across all types and future steps, both early and late LP substantially outperform the raw-input baseline. The gains are most significant for the $Straight$ and $Reverse$ cases, and performance gradually improves, then saturates as the number of scenes increases. At the same time, accuracy consistently declines as the future horizon lengthens.
The figure \ref{fig:other-types} shows the corresponding results for surrounding-vehicle probing. Here, accuracies are overall lower than in the ego case, and the gap between early and late LP is smaller and less stable, especially for $Turn$ and $Reverse$.

\begin{figure*}[h]
    \centering
    \includegraphics[width=0.9\linewidth]{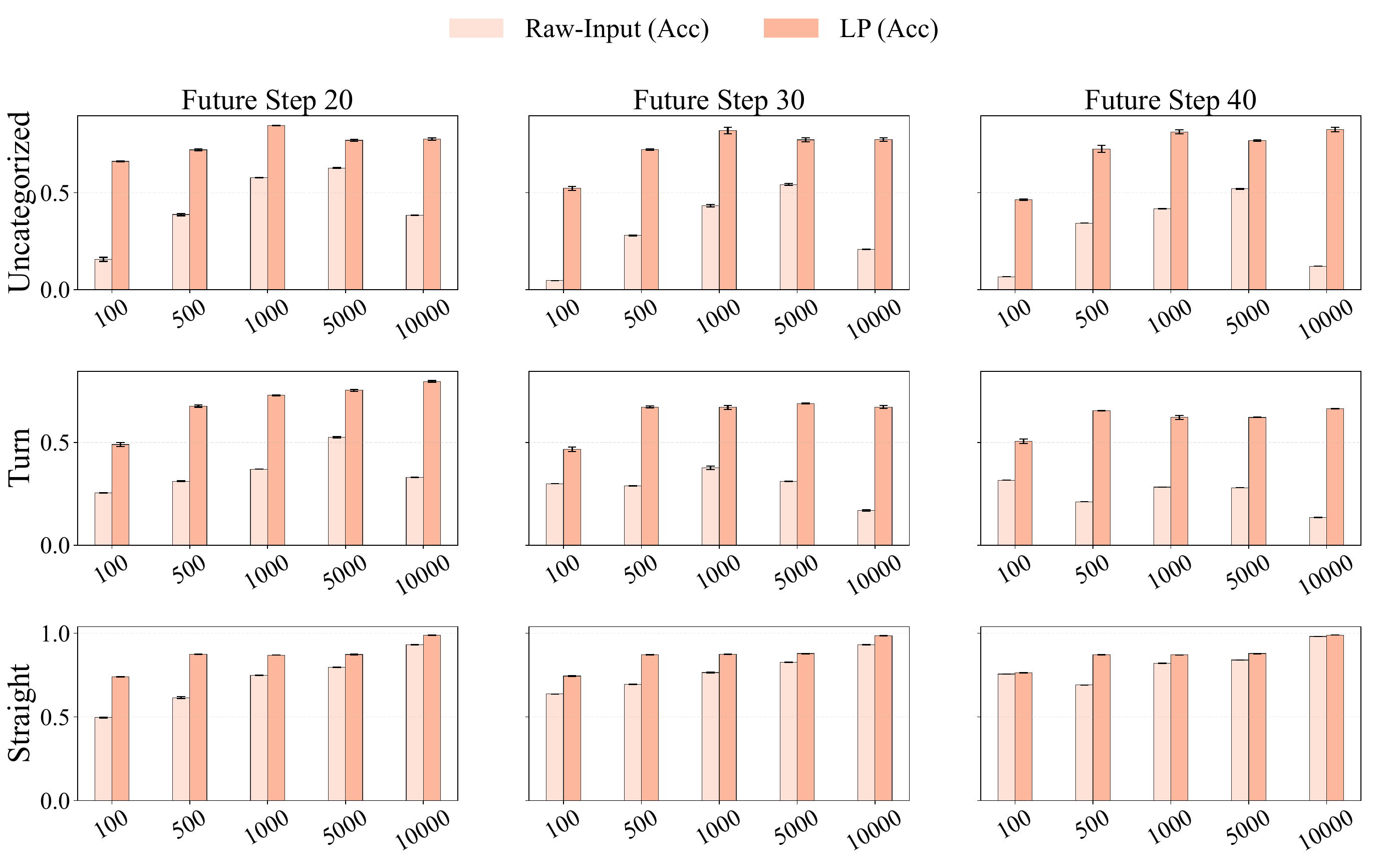}
    \caption{\textbf{Type accuracy of other future steps (20, 30, 40 (Ego) (RL)}: The row is type ($Normal$, $Turn$, $Straight$, and $Reverse$) and the column is future step (20, 30, 40).}
    \label{fig:ego-types-rl}
\end{figure*}
\newpage
\clearpage

%% file: Tables/9_lp_ego_setting.tex
\begin{table}[!htbp]
\caption{Linear Probing Setting Details for Ego Prediction.}
\label{tab:probing-setting-ego}
\centering
\resizebox{0.75\columnwidth}{!}{%
\begin{tabular}{c|c|c|c|c|c}
\hline
\textbf{\#Scenes} & \makecell{\textbf{Gradient}\\\textbf{(Step)}} & \makecell{\textbf{\#Samples}\\\textbf{(fs@10)}} & \makecell{\textbf{\#Samples}\\\textbf{(fs@20)}} & \makecell{\textbf{\#Samples}\\\textbf{(fs@30)}} & \makecell{\textbf{\#Samples}\\\textbf{(fs@40)}}\\
\hline
100    & 5K    & 37K   & 31K   & 26K   & 20K \\
500    & 12K   & 181K  & 152K  & 125K  & 99K \\
1000   & 20K   & 355K  & 297K  & 244K  & 193K \\
5000   & 50K   & 1.7M  & 1.4M  & 1.2M  & 937K \\
10,000 & 75K   & 3.5M  & 2.9M  & 2.3M  & 1.9M \\
20,000 & 100K  & 7M    & 5.8M  & 4.8M  & 3.8M \\
40,000 & 125K  & 13.9M & 11.6M & 9.5M  & 756K \\
80,000 & 150K  & 27.7M & 23.2M & 19M   & 15.1M \\
\hline
\end{tabular}%
}
\end{table}

%% file: Tables/5_f1_macro_ego.tex
\begin{table}[!htbp]
\centering
\caption{F1 Macro differences across future steps. (a) BC, (b) RL. $Info\ Loss$ is (Early LP $-$ Late LP), and $LP - Raw$ is the difference between trained model probing and the raw-input model.}
\label{tab:f1_diff_ego_column}

\begin{subtable}{\columnwidth}
\centering
\caption{BC}
\label{tab:f1_diff_ego_bc}
\resizebox{\linewidth}{!}{
\begin{tabular}{lrrrrrrrr}
\toprule
{} & \multicolumn{2}{c}{FS=10} & \multicolumn{2}{c}{FS=20} & \multicolumn{2}{c}{FS=30} & \multicolumn{2}{c}{FS=40} \\
{} & Info Loss. & LP - Raw & Info Loss. & LP - Raw & Info Loss. & LP - Raw & Info Loss. & LP - Raw \\
\midrule
100 & 0.0060 & 0.0847 & 0.0186 & 0.1522 & 0.0241 & 0.1849 & 0.0312 & 0.1784 \\
500 & 0.0086 & 0.1652 & 0.0150 & 0.2425 & 0.0226 & 0.2255 & 0.0330 & 0.2307 \\
1000 & 0.0047 & 0.1502 & 0.0122 & 0.2339 & 0.0262 & 0.2167 & 0.0424 & 0.2068 \\
5000 & -0.0078 & 0.1659 & 0.0012 & 0.2190 & 0.0130 & 0.1992 & 0.0251 & 0.1946 \\
10000 & -0.0109 & 0.1615 & -0.0030 & 0.2218 & 0.0082 & 0.1936 & 0.0210 & 0.1888 \\
20000 & -0.0192 & 0.1731 & -0.0113 & 0.2188 & 0.0026 & 0.1930 & 0.0158 & 0.1925 \\
40000 & -0.0190 & 0.1674 & -0.0100 & 0.2063 & 0.0032 & 0.1887 & 0.0186 & 0.1799 \\
80000 & -0.0185 & 0.1701 & -0.0121 & 0.2151 & -0.0004 & 0.1888 & 0.0143 & 0.1800 \\
\bottomrule
\end{tabular}
}
\end{subtable}

\begin{subtable}{\columnwidth}
\centering
\caption{RL}
\label{tab:f1_diff_ego_rl}
\resizebox{0.55\linewidth}{!}{
\begin{tabular}{lcccc}
\toprule
\#Scenes & FS=10 & FS=20 & FS=30 & FS=40 \\
\midrule
100 & 0.2207 & 0.3234 & 0.2487 & 0.1218 \\
500 & 0.2872 & 0.3494 & 0.2616 & 0.2799 \\
1000 & 0.1998 & 0.2476 & 0.2240 & 0.1567 \\
5000 & 0.1422 & 0.1859 & 0.1823 & 0.1801 \\
10000 & 0.2893 & 0.3191 & 0.3394 & 0.2771 \\
\bottomrule
\end{tabular}
}
\end{subtable}
\end{table}

%% file: Tables/6_info_loss_ego.tex
\begin{table}[!htbp]
\centering
\caption{Accuracy differences by action types. (a) BC at future step 10, (b) RL at future step 10. $Info\ Loss$ is (Early LP $-$ Late LP), and $LP - Raw$ is the difference between model probing and the raw-input model.}
\label{tab:acc_diff_types_ego}

\begin{subtable}{\columnwidth}
\centering
\caption{BC (Future step = 10)}
\label{tab:acc_diff_types_bc_ego}
\resizebox{\linewidth}{!}{
\begin{tabular}{lrrrrrrrr}
\toprule
{} & \multicolumn{2}{c}{Normal} & \multicolumn{2}{c}{Turn} & \multicolumn{2}{c}{Straight} & \multicolumn{2}{c}{Reverse} \\
{} & Info Loss. & LP - Raw & Info Loss. & LP - Raw & Info Loss. & LP - Raw & Info Loss. & LP - Raw \\
\midrule
100 & 0.0109 & 0.1016 & -0.0056 & 0.0196 & 0.0113 & 0.0717 & 0.0610 & 0.1219 \\
500 & 0.0141 & 0.1666 & -0.0036 & 0.0431 & 0.0172 & 0.1589 & -0.2088 & 0.5369 \\
1000 & 0.0054 & 0.1454 & -0.0031 & 0.0314 & 0.0022 & 0.1296 & 0.0113 & 0.5561 \\
5000 & -0.0047 & 0.1505 & -0.0102 & 0.0488 & 0.0001 & 0.1492 & -0.0115 & 0.6478 \\
10000 & -0.0090 & 0.1479 & -0.0130 & 0.0476 & -0.0077 & 0.1411 & -0.0365 & 0.6517 \\
20000 & -0.0159 & 0.1554 & -0.0156 & 0.0534 & -0.0132 & 0.1563 & -0.0049 & 0.6392 \\
40000 & -0.0136 & 0.1517 & -0.0155 & 0.0541 & -0.0142 & 0.1423 & -0.0097 & 0.6406 \\
80000 & -0.0145 & 0.1556 & -0.0141 & 0.0527 & -0.0115 & 0.1477 & 0.0166 & 0.6605 \\
\bottomrule
\end{tabular}
}
\end{subtable}

\begin{subtable}{\columnwidth}
\centering
\caption{RL (Future step = 10)}
\label{tab:acc_diff_types_ego_ego}
\resizebox{0.55\linewidth}{!}{
\begin{tabular}{lrrrr}
\toprule
{} & Normal & Turn & Straight & Reverse \\
{} & IL - Raw & IL - Raw & IL - Raw & IL - Raw \\
\midrule
100 & 0.0712 & 0.1789 & 0.2450 & -- \\
500 & 0.1996 & 0.2314 & 0.1587 & -- \\
1000 & 0.1269 & 0.0982 & 0.1205 & -- \\
5000 & 0.0353 & 0.0844 & 0.0604 & -- \\
10000 & 0.1562 & 0.3351 & 0.0744 & -- \\
\bottomrule
\end{tabular}
}
\end{subtable}
\end{table}

%% file: appendix/f_further.tex
\section{Further Analysis}

\subsection{Ego-linear Probing Visualization Comparison.} 
\label{app:perturbed}
\begin{figure*}[h]
    \centering
    \includegraphics[width=0.95\linewidth]{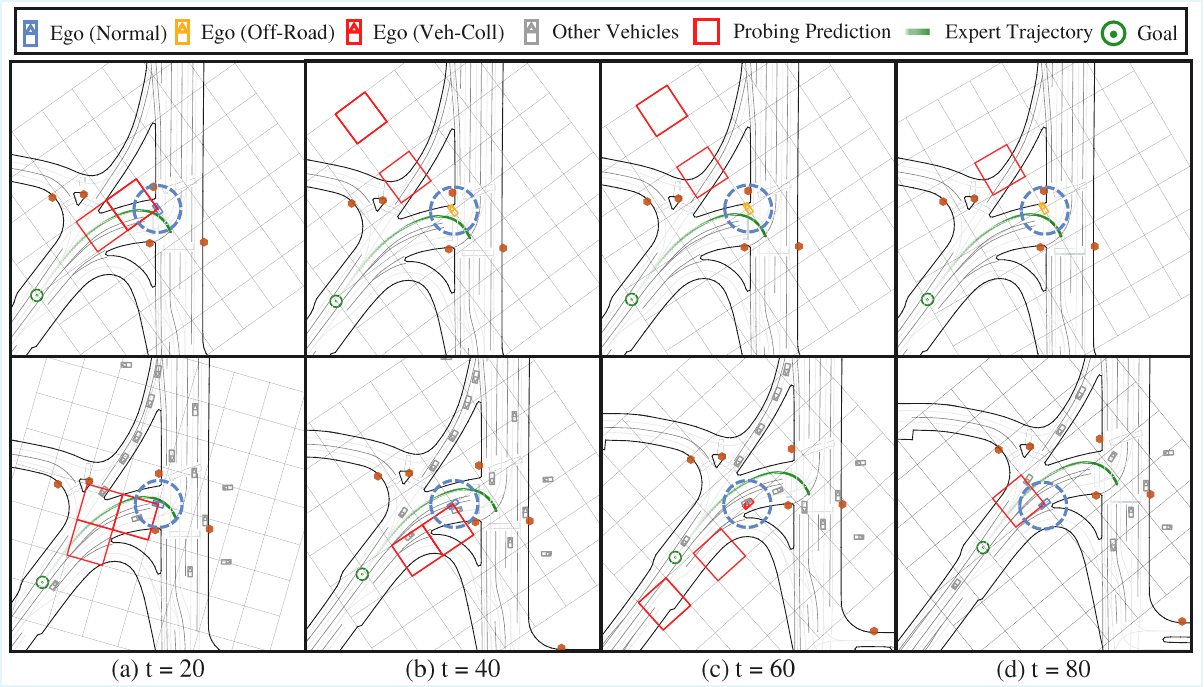}
    \caption{\textbf{Comparison of ego-linear probing in single AV driving and all vehicles driving with rendered examples}: The visualization of ego-linear probing when single AV drives \textbf{(Top)} and all vehicles include AV drive \textbf{(Bottom)}. The expert trajectory is shown as a dark green line. The ego AV is blue; it turns yellow when off-road and red when a vehicle collides.}
    \label{fig:further-probing}
\end{figure*}
To better understand the degradation of single AV driving performance, we visualize ego-linear probing in simulation. We compare ego-linear probing results across the 100 scenes with the largest performance gap. As shown in Figure \ref{fig:further-probing}, the top row shows the single AV driving, and the bottom row shows all vehicles driving. In the single AV case, the ego collides around $t=40$ and subsequently loses the ability to plan toward the goal. By contrast, in the all-vehicles-driving case, the model plans slightly below the ground truth path at $t=40$, leading to a collision around $t=60$; however, even after the crash, the AV continues to plan accurately toward the goal and recovers the exact trajectory at $t=80$. From the results, we suspect the cause of this situation is the collapse of ego-linear probing, which also provides strong evidence that the presence of surrounding vehicles affects ego AV planning, given the dataset's lack of single-AV driving.
\newpage
\subsection{Additional Near-Collision Analysis}
\label{app:near-collision}
\begin{figure*}[h]
    \centering
    \includegraphics[width=0.95\linewidth]{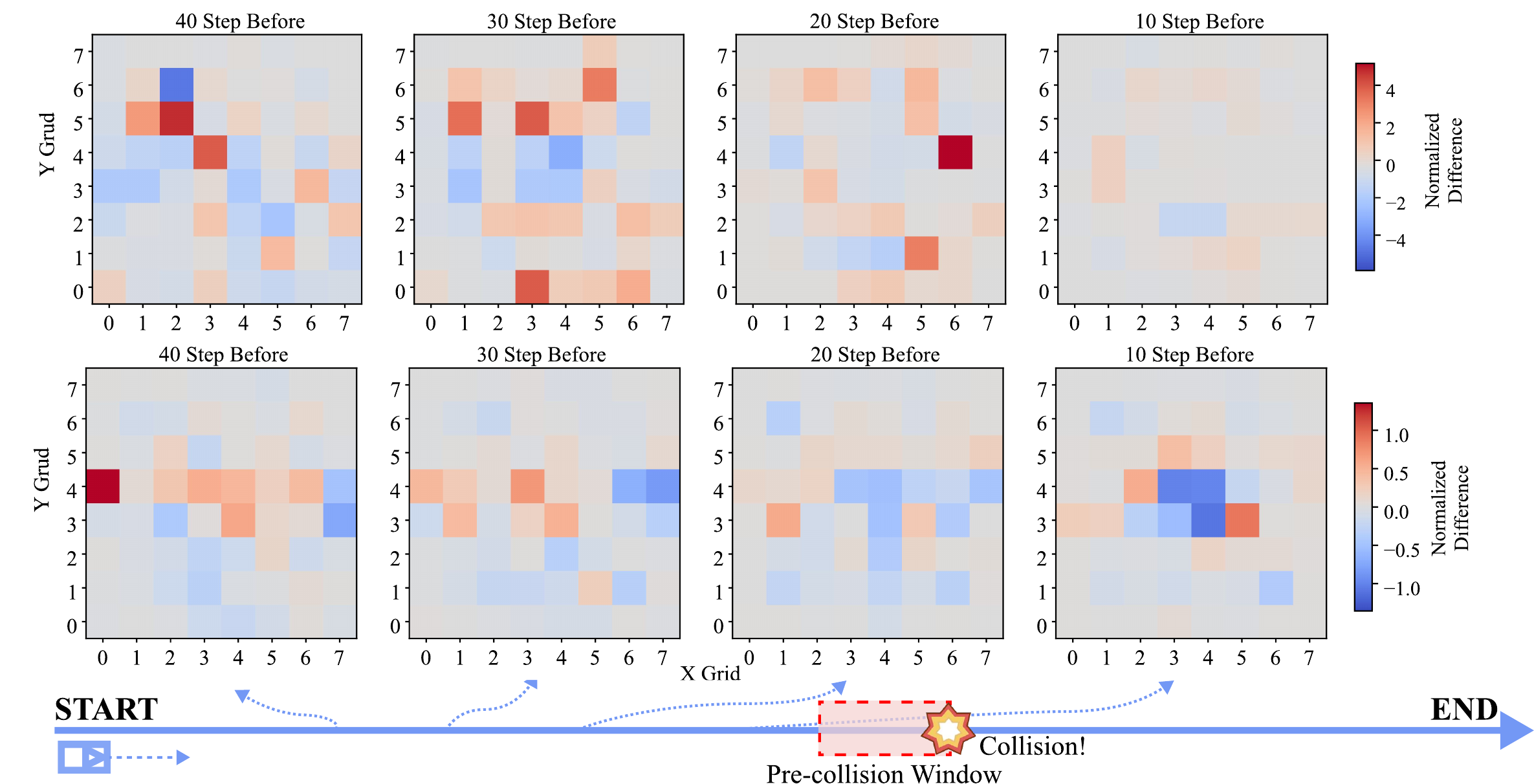}
    \caption{\textbf{Near-collision Analysis. (Off-road)}
    Each heatmap shows the normalized difference between the predicted probability in the pre-collision window $w$ and that of each spatial grid over the full episode, computed as $\frac{w-a_{\mathrm{grid}}}{a}$, at 10 to 40 steps before collision. 
    The red dot marks the ego vehicle position. Dark gray cells indicate grids that were filtered out due to insufficient collision samples. Top: RL, Bottom: BC}
    \label{fig:off-road}
\end{figure*}

\newpage
\clearpage

%% file: appendix/g_intervention.tex
\section{Intervention Results}
\label{app:intervention}
\input{Tables/10_intervention_stat}
We report the detailed results of the intervention experiment in Table \ref{tab:intervention-label}. 
For BC, most \textit{route change} cases are successfully modified to avoid collisions, but interventions requiring increased speed remain difficult, with only 4/15 successful cases. 
In contrast, RL and SMART (IL) show stronger performance in \textit{increase speed} cases, achieving 11/16 and 13/18 successes, respectively. 
IL also performs well in \textit{slow speed} cases, with 8/9 successful interventions.

Overall, the ego-planning probe changes in 36/59 BC cases, 37/58 RL cases, and 40/64 IL cases. These results suggest that the proposed intervention framework applies not only to BC but also to RL and IL models, and that surrounding-vehicle predictions can causally influence ego planning across different driving models. 

\subsection{Failure Case}
\begin{figure*}[h]
    \centering
    \includegraphics[width=0.95\linewidth]{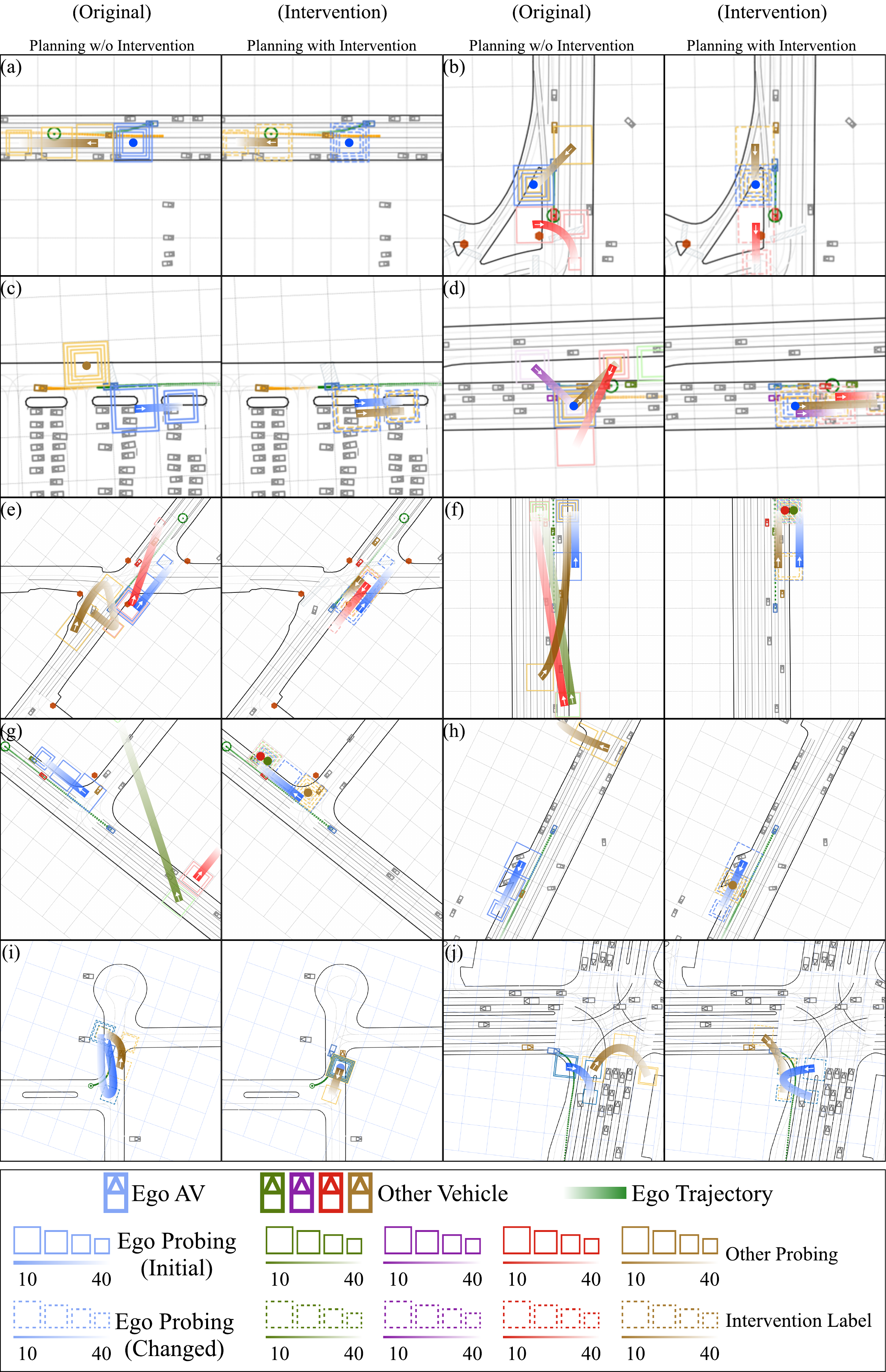}
    \caption{\textbf{Failure cases of intervention}: The visualization examples for failure cases of changing labels. (a)-(d): BC, (e)-(h): RL, (i)-(j): IL}
    \label{fig:failure cases}
\end{figure*}

We analyze the failure cases, as illustrated in Figure \ref{fig:failure cases}, where there is no change compared to the initial prediction of ego planning. Interestingly, most cases are increasing in speed. As in cases (a) and (d), we make the front vehicles go faster so that the ego vehicle can increase its speed. However, the BC model remains on the same grid as before. Especially in case (d), even if we move multiple vehicles that could affect the ego's future position, the BC model remains in the same grid. In the case of (b) and (c), we increase the following vehicles' speed so that the ego prediction should speed up. However, there is no change in the BC model. In contrast, the RL policies have difficulty changing lanes, as seen in (e), (f), and (g). In (h), we expect the RL policy to increase speed, but it fails to do so. Lastly, in the IL model, as shown in (i), the ego vehicle remains in the grid before intervention, and the other vehicles' predictions are completely wrong. We restore the other vehicle to move in the correct direction, but the ego planner changes direction incorrectly. The same situation occurs in (j); we correct the vehicle prediction, but suddenly, ego planning goes the wrong way. 

\begin{figure*}[h]
    \centering
    \includegraphics[width=0.82\linewidth]{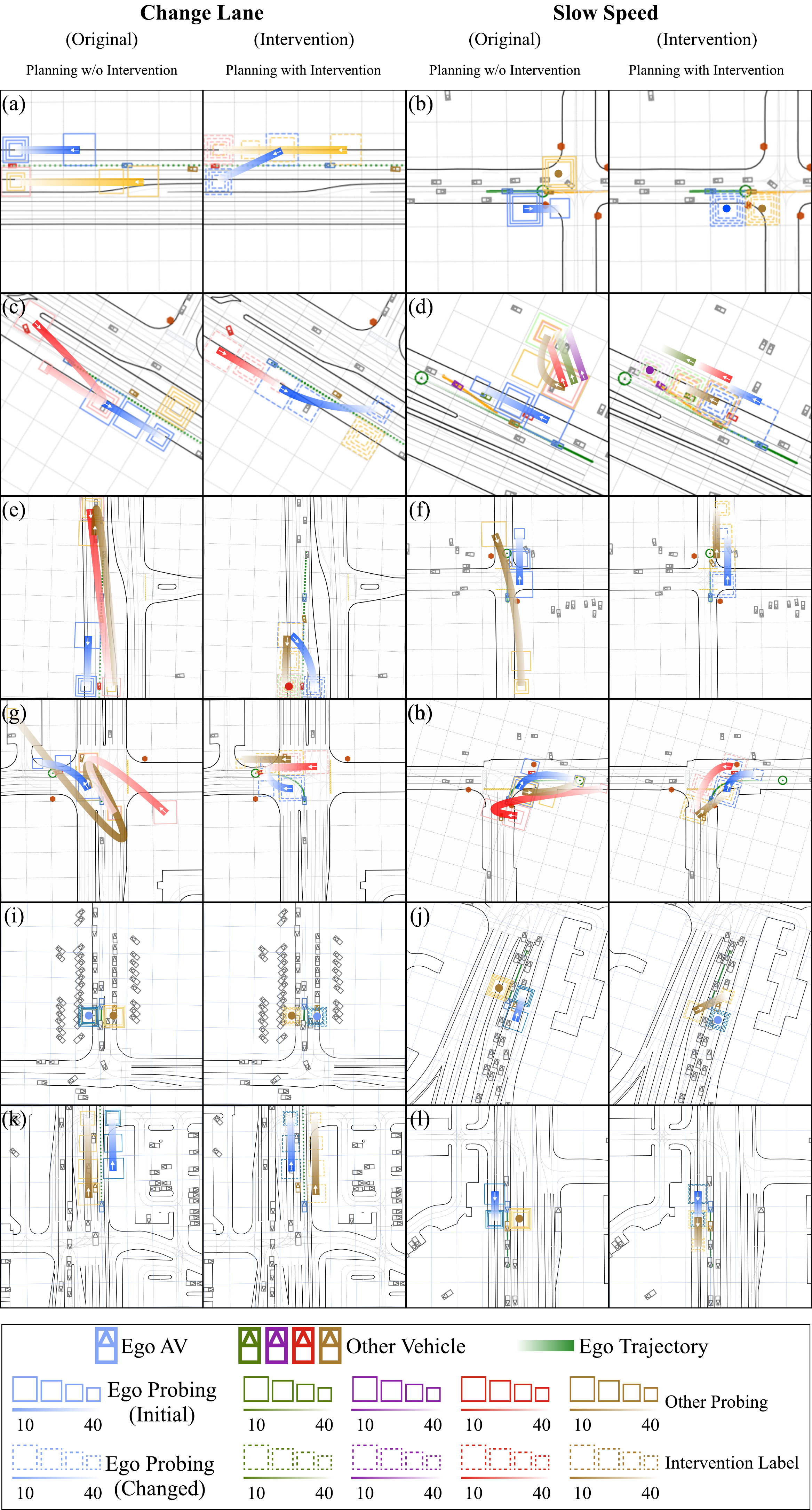}
    \caption{\textbf{Additional Examples for Adaptiveness cases}: The visualization examples for adaptiveness cases. \textbf{Change Lane}: ego linear probing changes the lane to avoid collision. \textbf{Slowing Speed}: When we disturb the trajectory, the model slows down the speed. (a)-(d): BC, (e)-(h): RL, (i)-(l): IL}
    \label{fig:adaptiveness}
\end{figure*}

\subsection{Additional Intervention Cases}
As shown in Figure \ref{fig:adaptiveness}, panels (a)--(d) present additional BC adaptiveness cases, while panels (e)--(h) present RL adaptiveness cases.
For the BC cases, panels (a) and (c) illustrate change lane interventions. In (a), the intervention induces a lane change through a single interaction with the yellow vehicle. In (c), the red vehicle follows from behind while the yellow vehicle changes lanes, requiring the ego to maintain speed and change lanes as well; the BC model successfully performs this behavior. Panels (b) and (d) show slowdown interventions. In (b), when the yellow lead vehicle is kept in the ego's future grid cells, the ego remains behind it and slows its progress. In (d), we place multiple surrounding-vehicle trajectories in front of the ego, and the BC model responds by slowing down accordingly.

For the RL cases, panels (e) and (g) show change lane interventions, whereas panels (f) and (h) show slowdown interventions. In (e), the yellow and red vehicles originally occupy implausible positions, but when they are placed on the ego's predicted path, the ego changes its trajectory to avoid them. In (f) and (h), we modify previously implausible surrounding-vehicle trajectories to make them more feasible, and the RL model correspondingly reduces speed. Finally, in (g), after aligning an initially unrealistic trajectory, the ego's predicted motion is adjusted so that it no longer overlaps in time with the intervened vehicle.

For the IL cases, panels (i) and (k) show that the ego vehicle successfully adjusts its planning when we intervene in the surrounding-vehicle prediction to align with the original planning. In the cases of (j) and (l), we intervene in the ego's planning to stay on a further trajectory, so we expect the ego to change plans to slow down. The ego plans successfully to slow down to avoid a collision.

\subsection{Additional Recovery Case}
\begin{figure*}[h]
    \centering
    \includegraphics[width=0.75\linewidth]{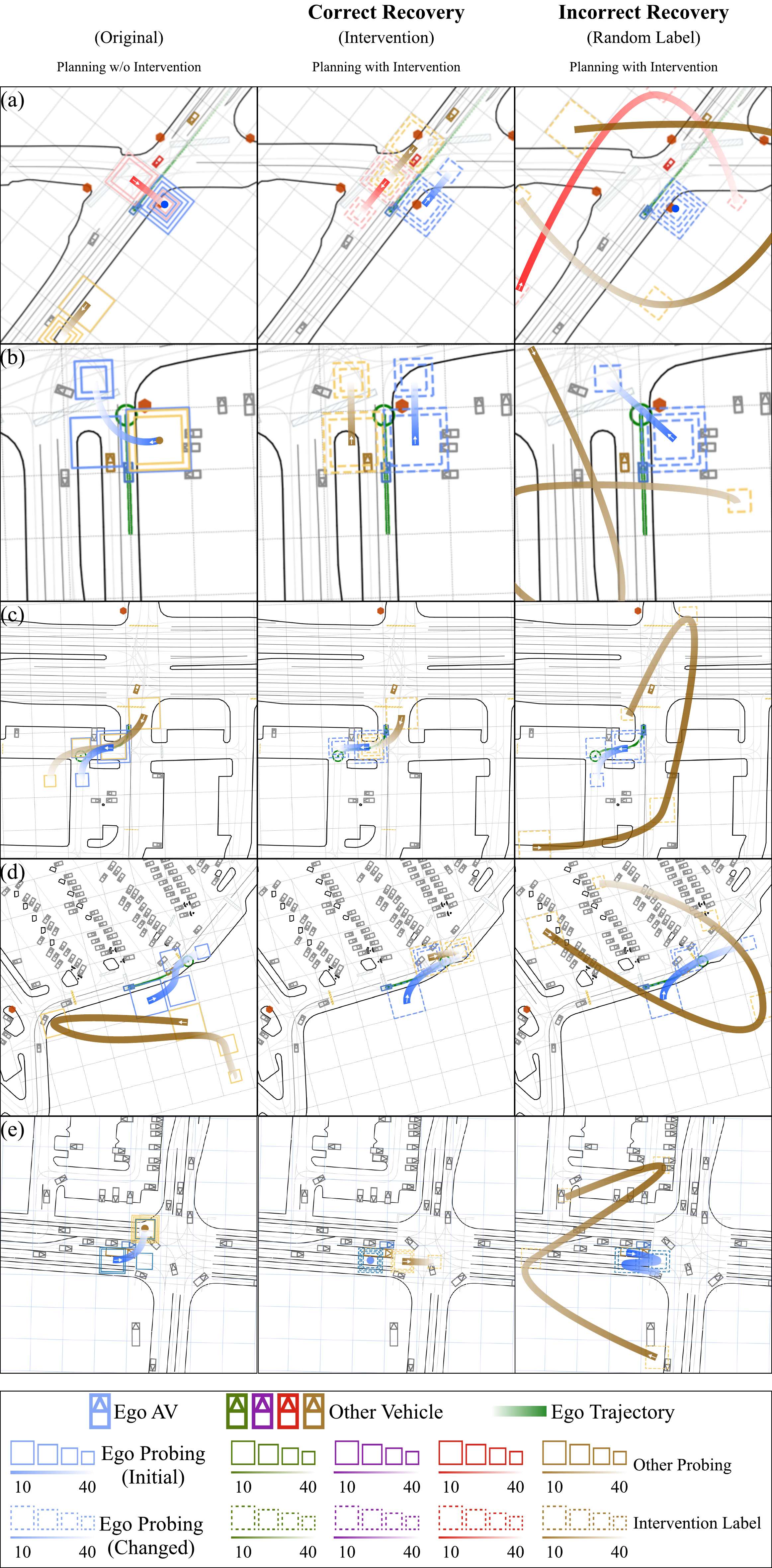}
    \caption{\textbf{Additional Examples for Recovery cases}: The visualization examples for recovery cases. \textbf{Correct Recovery}: When we change the label to align with the correct expert trajectory of the surrounding vehicle. \textbf{Incorrect Recovery}: When we set the label to a random label. (a)-(b): BC, (c)-(d): RL, (e): IL}
    \label{fig:recovery}
\end{figure*}

As shown in Figure \ref{fig:recovery}, we describe the additional results of recovery cases. To evaluate the correction of the recovery label, we visualize the correctly recovered cases, which we label based on the surrounding vehicle trajectories and the random labels of other vehicles. In panel (a), the original prediction of the yellow vehicle is totally wrong, and the red vehicle is slightly wrong at 30 and 40 steps after. Therefore, we recover the red and yellow vehicle predictions to straighten and verify that the ego vehicle increases its speed slightly faster than before. Interestingly, when we assign a random label, the ego-vehicle prediction is the same as the initial prediction. Panel (b) examples show that the other vehicle prediction overlaps, so the ego prediction needs to avoid other vehicles. Therefore, we change the other vehicle's prediction along a straight line and show that the ego prediction changes in the same way, whereas random labeling tends to stay in the same grid for longer. In panel (c), which depicts the RL model's intervention, we recover the other vehicle's trajectory to follow the ego vehicle. Then, the ego vehicle adjusts to stay at the destination and avoid overwhelming it, since it knows the yellow vehicle will not overtake the ego vehicle while the random label has the same planning as before the intervention. In panel (d), the RL completely mispredicts the vehicle in front. When we recover that it is feasible, the model did not predict the off-road area in 20 timesteps after. However, the random label indicates the wrong destination after 40 timesteps. Lastly, in (e), we recover the ego's trajectory as we approach a nearby vehicle to go straight. However, when we intervene randomly, the ego plan becomes corrupted and zigzags.

\newpage
\clearpage

%% file: Tables/10_intervention_stat.tex
\begin{table}[h]
  \caption{Statistics of intervention cases}
  \label{tab:intervention-label}
  \centering
  \renewcommand{\arraystretch}{1.1}

  \begin{subtable}[t]{0.47\columnwidth}
    \caption{BC: Success-rate cases}
    \centering
    \begin{tabular}{@{}lrrr@{}}
      \toprule
      Case & Success & Fail & Total \\
      \midrule
      route change   & 20 &  3 & 23 \\ 
      increase speed &  4 & 11 & 15  \\ 
      slow speed     &  4 &  1 &  5 \\  
      \bottomrule
    \end{tabular}
  \end{subtable}
  \hfill
  \begin{subtable}[t]{0.47\columnwidth}
    \caption{BC: Adaptiveness and recovery}
    \centering
    \begin{tabular}{@{}lrrr@{}}
      \toprule
      Case & Success & Fail & Total \\
      \midrule
      Adaptiveness & 28 & 15 & 43 \\
      Recovery     &  8 &  8 & 16 \\
      \bottomrule
    \end{tabular}
  \end{subtable}

  \vspace{2mm}

  \begin{subtable}[t]{0.47\columnwidth}
    \caption{RL: Success-rate cases}
    \centering
    \begin{tabular}{@{}lrrr@{}}
      \toprule
      Case & Success & Fail & Total  \\
      \midrule
      route change   & 12 & 9 & 21 \\
      increase speed & 11 & 5 & 16 \\
      slow speed     & 9 & 7 & 16 \\
      \bottomrule
    \end{tabular}
  \end{subtable}
  \hfill
  \begin{subtable}[t]{0.47\columnwidth}
    \caption{RL: Adaptiveness and recovery}
    \centering
    \begin{tabular}{@{}lrrr@{}}
      \toprule
      Case & Success & Fail & Total\\
      \midrule
      Adaptiveness & 32 & 21 & 53 \\
      Recovery     & 5 & 0 & 5 \\
      \bottomrule
    \end{tabular}
  \end{subtable}

  \vspace{2mm}

  \begin{subtable}[t]{0.47\columnwidth}
    \caption{SMART (IL): Success-rate cases}
    \centering
    \begin{tabular}{@{}lrrr@{}}
      \toprule
      Case & Success & Fail & Total \\
      \midrule
      route change   & 13 & 10 & 23 \\
      increase speed & 13 &  5 & 18 \\
      slow speed     &  8 &  1 &  9 \\
      \bottomrule
    \end{tabular}
  \end{subtable}
  \hfill
  \begin{subtable}[t]{0.47\columnwidth}
    \caption{SMART (IL): Adaptiveness and recovery}
    \centering
    \begin{tabular}{@{}lrrr@{}}
      \toprule
      Case & Success & Fail & Total \\
      \midrule
      Adaptiveness & 33 & 18 & 51 \\
      Recovery     &  7 &  6 & 13 \\
      \bottomrule
    \end{tabular}
  \end{subtable}

\end{table}

%% file: neurips2026/checklist.tex
\section*{NeurIPS Paper Checklist}

\begin{enumerate}

\item {\bf Claims}
    \item[] Question: Do the main claims made in the abstract and introduction accurately reflect the paper's contributions and scope?
    \item[] Answer: \answerYes{} 
    \item[] Justification: The abstract and introduction accurately reflect the paper’s scope and contributions. The paper analyzes how autonomous driving models represent and use information about surrounding vehicles, using linear probing, perturbed closed-loop simulations, near-collision analysis, and intervention experiments. These analyses support the central claims that prediction and planning capabilities emerge with scale, that nominal closed-loop success can mask weaknesses in surrounding-vehicle reasoning, and that improving internal predictions can causally improve ego planning. The experimental sections are well aligned with these stated goals.
    \item[] Guidelines:
    \begin{itemize}
        \item The answer \answerNA{} means that the abstract and introduction do not include the claims made in the paper.
        \item The abstract and/or introduction should clearly state the claims made, including the contributions made in the paper and important assumptions and limitations. A \answerNo{} or \answerNA{} answer to this question will not be perceived well by the reviewers. 
        \item The claims made should match theoretical and experimental results, and reflect how much the results can be expected to generalize to other settings. 
        \item It is fine to include aspirational goals as motivation as long as it is clear that these goals are not attained by the paper. 
    \end{itemize}

\item {\bf Limitations}
    \item[] Question: Does the paper discuss the limitations of the work performed by the authors?
    \item[] Answer: \answerYes{} 
    \item[] Justification: We discussed the limitations of the work and future works in Section 4.
    \item[] Guidelines:
    \begin{itemize}
        \item The answer \answerNA{} means that the paper has no limitation while the answer \answerNo{} means that the paper has limitations, but those are not discussed in the paper. 
        \item The authors are encouraged to create a separate ``Limitations'' section in their paper.
        \item The paper should point out any strong assumptions and how robust the results are to violations of these assumptions (e.g., independence assumptions, noiseless settings, model well-specification, asymptotic approximations only holding locally). The authors should reflect on how these assumptions might be violated in practice and what the implications would be.
        \item The authors should reflect on the scope of the claims made, e.g., if the approach was only tested on a few datasets or with a few runs. In general, empirical results often depend on implicit assumptions, which should be articulated.
        \item The authors should reflect on the factors that influence the performance of the approach. For example, a facial recognition algorithm may perform poorly when image resolution is low or images are taken in low lighting. Or a speech-to-text system might not be used reliably to provide closed captions for online lectures because it fails to handle technical jargon.
        \item The authors should discuss the computational efficiency of the proposed algorithms and how they scale with dataset size.
        \item If applicable, the authors should discuss possible limitations of their approach to address problems of privacy and fairness.
        \item While the authors might fear that complete honesty about limitations might be used by reviewers as grounds for rejection, a worse outcome might be that reviewers discover limitations that aren't acknowledged in the paper. The authors should use their best judgment and recognize that individual actions in favor of transparency play an important role in developing norms that preserve the integrity of the community. Reviewers will be specifically instructed to not penalize honesty concerning limitations.
    \end{itemize}

\item {\bf Theory assumptions and proofs}
    \item[] Question: For each theoretical result, does the paper provide the full set of assumptions and a complete (and correct) proof?
    \item[] Answer: \answerNA{} 
    \item[] Justification: In this paper, we do not propose new theorems. In section 3.2, we conduct the scaling-law experiment and show that the driving models follow this scale law.
    \item[] Guidelines:
    \begin{itemize}
        \item The answer \answerNA{} means that the paper does not include theoretical results. 
        \item All the theorems, formulas, and proofs in the paper should be numbered and cross-referenced.
        \item All assumptions should be clearly stated or referenced in the statement of any theorems.
        \item The proofs can either appear in the main paper or the supplemental material, but if they appear in the supplemental material, the authors are encouraged to provide a short proof sketch to provide intuition. 
        \item Inversely, any informal proof provided in the core of the paper should be complemented by formal proofs provided in appendix or supplemental material.
        \item Theorems and Lemmas that the proof relies upon should be properly referenced. 
    \end{itemize}

    \item {\bf Experimental result reproducibility}
    \item[] Question: Does the paper fully disclose all the information needed to reproduce the main experimental results of the paper to the extent that it affects the main claims and/or conclusions of the paper (regardless of whether the code and data are provided or not)?
    \item[] Answer: \answerYes{} 
    \item[] Justification: This paper provides model settings in Appendix B and experiment settings in Appendix C.1, D.1, and E.1.
    \item[] Guidelines:
    \begin{itemize}
        \item The answer \answerNA{} means that the paper does not include experiments.
        \item If the paper includes experiments, a \answerNo{} answer to this question will not be perceived well by the reviewers: Making the paper reproducible is important, regardless of whether the code and data are provided or not.
        \item If the contribution is a dataset and\slash or model, the authors should describe the steps taken to make their results reproducible or verifiable. 
        \item Depending on the contribution, reproducibility can be accomplished in various ways. For example, if the contribution is a novel architecture, describing the architecture fully might suffice, or if the contribution is a specific model and empirical evaluation, it may be necessary to either make it possible for others to replicate the model with the same dataset, or provide access to the model. In general. releasing code and data is often one good way to accomplish this, but reproducibility can also be provided via detailed instructions for how to replicate the results, access to a hosted model (e.g., in the case of a large language model), releasing of a model checkpoint, or other means that are appropriate to the research performed.
        \item While NeurIPS does not require releasing code, the conference does require all submissions to provide some reasonable avenue for reproducibility, which may depend on the nature of the contribution. For example
        \begin{enumerate}
            \item If the contribution is primarily a new algorithm, the paper should make it clear how to reproduce that algorithm.
            \item If the contribution is primarily a new model architecture, the paper should describe the architecture clearly and fully.
            \item If the contribution is a new model (e.g., a large language model), then there should either be a way to access this model for reproducing the results or a way to reproduce the model (e.g., with an open-source dataset or instructions for how to construct the dataset).
            \item We recognize that reproducibility may be tricky in some cases, in which case authors are welcome to describe the particular way they provide for reproducibility. In the case of closed-source models, it may be that access to the model is limited in some way (e.g., to registered users), but it should be possible for other researchers to have some path to reproducing or verifying the results.
        \end{enumerate}
    \end{itemize}

\item {\bf Open access to data and code}
    \item[] Question: Does the paper provide open access to the data and code, with sufficient instructions to faithfully reproduce the main experimental results, as described in supplemental material?
    \item[] Answer: \answerNo{} 
    \item[] Justification: Although the data and code will not accompany in the submission version,
they will be released upon publication.
    \item[] Guidelines:
    \begin{itemize}
        \item The answer \answerNA{} means that paper does not include experiments requiring code.
        \item Please see the NeurIPS code and data submission guidelines (\url{https://neurips.cc/public/guides/CodeSubmissionPolicy}) for more details.
        \item While we encourage the release of code and data, we understand that this might not be possible, so \answerNo{} is an acceptable answer. Papers cannot be rejected simply for not including code, unless this is central to the contribution (e.g., for a new open-source benchmark).
        \item The instructions should contain the exact command and environment needed to run to reproduce the results. See the NeurIPS code and data submission guidelines (\url{https://neurips.cc/public/guides/CodeSubmissionPolicy}) for more details.
        \item The authors should provide instructions on data access and preparation, including how to access the raw data, preprocessed data, intermediate data, and generated data, etc.
        \item The authors should provide scripts to reproduce all experimental results for the new proposed method and baselines. If only a subset of experiments are reproducible, they should state which ones are omitted from the script and why.
        \item At submission time, to preserve anonymity, the authors should release anonymized versions (if applicable).
        \item Providing as much information as possible in supplemental material (appended to the paper) is recommended, but including URLs to data and code is permitted.
    \end{itemize}

\item {\bf Experimental setting/details}
    \item[] Question: Does the paper specify all the training and test details (e.g., data splits, hyperparameters, how they were chosen, type of optimizer) necessary to understand the results?
    \item[] Answer: \answerYes{} 
    \item[] Justification: We briefly introduce the experiment settings in Section 3.1. Detailed settings are in the Appendices.
    \item[] Guidelines:
    \begin{itemize}
        \item The answer \answerNA{} means that the paper does not include experiments.
        \item The experimental setting should be presented in the core of the paper to a level of detail that is necessary to appreciate the results and make sense of them.
        \item The full details can be provided either with the code, in appendix, or as supplemental material.
    \end{itemize}

\item {\bf Experiment statistical significance}
    \item[] Question: Does the paper report error bars suitably and correctly defined or other appropriate information about the statistical significance of the experiments?
    \item[] Answer: \answerYes{} 
    \item[] Justification: We include the error bar and standard deviation in all experiments.
    \item[] Guidelines:
    \begin{itemize}
        \item The answer \answerNA{} means that the paper does not include experiments.
        \item The authors should answer \answerYes{} if the results are accompanied by error bars, confidence intervals, or statistical significance tests, at least for the experiments that support the main claims of the paper.
        \item The factors of variability that the error bars are capturing should be clearly stated (for example, train/test split, initialization, random drawing of some parameter, or overall run with given experimental conditions).
        \item The method for calculating the error bars should be explained (closed form formula, call to a library function, bootstrap, etc.)
        \item The assumptions made should be given (e.g., Normally distributed errors).
        \item It should be clear whether the error bar is the standard deviation or the standard error of the mean.
        \item It is OK to report 1-sigma error bars, but one should state it. The authors should preferably report a 2-sigma error bar than state that they have a 96\% CI, if the hypothesis of Normality of errors is not verified.
        \item For asymmetric distributions, the authors should be careful not to show in tables or figures symmetric error bars that would yield results that are out of range (e.g., negative error rates).
        \item If error bars are reported in tables or plots, the authors should explain in the text how they were calculated and reference the corresponding figures or tables in the text.
    \end{itemize}

\item {\bf Experiments compute resources}
    \item[] Question: For each experiment, does the paper provide sufficient information on the computer resources (type of compute workers, memory, time of execution) needed to reproduce the experiments?
    \item[] Answer: \answerYes{} 
    \item[] Justification: In Appendix A.2 and A.3, we recorded the computing resources of our main experiment.
    \item[] Guidelines:
    \begin{itemize}
        \item The answer \answerNA{} means that the paper does not include experiments.
        \item The paper should indicate the type of compute workers CPU or GPU, internal cluster, or cloud provider, including relevant memory and storage.
        \item The paper should provide the amount of compute required for each of the individual experimental runs as well as estimate the total compute. 
        \item The paper should disclose whether the full research project required more compute than the experiments reported in the paper (e.g., preliminary or failed experiments that didn't make it into the paper). 
    \end{itemize}
    
\item {\bf Code of ethics}
    \item[] Question: Does the research conducted in the paper conform, in every respect, with the NeurIPS Code of Ethics \url{https://neurips.cc/public/EthicsGuidelines}?
    \item[] Answer: \answerYes{} 
    \item[] Justification: 
    \item[] Guidelines:
    \begin{itemize}
        \item The answer \answerNA{} means that the authors have not reviewed the NeurIPS Code of Ethics.
        \item If the authors answer \answerNo, they should explain the special circumstances that require a deviation from the Code of Ethics.
        \item The authors should make sure to preserve anonymity (e.g., if there is a special consideration due to laws or regulations in their jurisdiction).
    \end{itemize}

\item {\bf Broader impacts}
    \item[] Question: Does the paper discuss both potential positive societal impacts and negative societal impacts of the work performed?
    \item[] Answer: \answerYes{} 
    \item[] Justification: We summarize the paper's contribution to the autonomous driving field. We analyzed the black-box deep learning autonomous driving models foucsed on utilization of surrounding vehicle information.
    \item[] Guidelines:
    \begin{itemize}
        \item The answer \answerNA{} means that there is no societal impact of the work performed.
        \item If the authors answer \answerNA{} or \answerNo, they should explain why their work has no societal impact or why the paper does not address societal impact.
        \item Examples of negative societal impacts include potential malicious or unintended uses (e.g., disinformation, generating fake profiles, surveillance), fairness considerations (e.g., deployment of technologies that could make decisions that unfairly impact specific groups), privacy considerations, and security considerations.
        \item The conference expects that many papers will be foundational research and not tied to particular applications, let alone deployments. However, if there is a direct path to any negative applications, the authors should point it out. For example, it is legitimate to point out that an improvement in the quality of generative models could be used to generate Deepfakes for disinformation. On the other hand, it is not needed to point out that a generic algorithm for optimizing neural networks could enable people to train models that generate Deepfakes faster.
        \item The authors should consider possible harms that could arise when the technology is being used as intended and functioning correctly, harms that could arise when the technology is being used as intended but gives incorrect results, and harms following from (intentional or unintentional) misuse of the technology.
        \item If there are negative societal impacts, the authors could also discuss possible mitigation strategies (e.g., gated release of models, providing defenses in addition to attacks, mechanisms for monitoring misuse, mechanisms to monitor how a system learns from feedback over time, improving the efficiency and accessibility of ML).
    \end{itemize}
    
\item {\bf Safeguards}
    \item[] Question: Does the paper describe safeguards that have been put in place for responsible release of data or models that have a high risk for misuse (e.g., pre-trained language models, image generators, or scraped datasets)?
    \item[] Answer: \answerNA{} 
    \item[] Justification: Our experiments were conducted on a simulation, not in the real world. 
    \item[] Guidelines:
    \begin{itemize}
        \item The answer \answerNA{} means that the paper poses no such risks.
        \item Released models that have a high risk for misuse or dual-use should be released with necessary safeguards to allow for controlled use of the model, for example by requiring that users adhere to usage guidelines or restrictions to access the model or implementing safety filters. 
        \item Datasets that have been scraped from the Internet could pose safety risks. The authors should describe how they avoided releasing unsafe images.
        \item We recognize that providing effective safeguards is challenging, and many papers do not require this, but we encourage authors to take this into account and make a best faith effort.
    \end{itemize}

\item {\bf Licenses for existing assets}
    \item[] Question: Are the creators or original owners of assets (e.g., code, data, models), used in the paper, properly credited and are the license and terms of use explicitly mentioned and properly respected?
    \item[] Answer: \answerYes{} 
    \item[] Justification: The datasets used in this paper are cited within this paper.
    \item[] Guidelines:
    \begin{itemize}
        \item The answer \answerNA{} means that the paper does not use existing assets.
        \item The authors should cite the original paper that produced the code package or dataset.
        \item The authors should state which version of the asset is used and, if possible, include a URL.
        \item The name of the license (e.g., CC-BY 4.0) should be included for each asset.
        \item For scraped data from a particular source (e.g., website), the copyright and terms of service of that source should be provided.
        \item If assets are released, the license, copyright information, and terms of use in the package should be provided. For popular datasets, \url{paperswithcode.com/datasets} has curated licenses for some datasets. Their licensing guide can help determine the license of a dataset.
        \item For existing datasets that are re-packaged, both the original license and the license of the derived asset (if it has changed) should be provided.
        \item If this information is not available online, the authors are encouraged to reach out to the asset's creators.
    \end{itemize}

\item {\bf New assets}
    \item[] Question: Are new assets introduced in the paper well documented and is the documentation provided alongside the assets?
    \item[] Answer: \answerYes{} 
    \item[] Justification:
    \item[] Guidelines:
    \begin{itemize}
        \item The answer \answerNA{} means that the paper does not release new assets.
        \item Researchers should communicate the details of the dataset\slash code\slash model as part of their submissions via structured templates. This includes details about training, license, limitations, etc. 
        \item The paper should discuss whether and how consent was obtained from people whose asset is used.
        \item At submission time, remember to anonymize your assets (if applicable). You can either create an anonymized URL or include an anonymized zip file.
    \end{itemize}

\item {\bf Crowdsourcing and research with human subjects}
    \item[] Question: For crowdsourcing experiments and research with human subjects, does the paper include the full text of instructions given to participants and screenshots, if applicable, as well as details about compensation (if any)? 
    \item[] Answer: \answerNA{} 
    \item[] Justification:
    \item[] Guidelines:
    \begin{itemize}
        \item The answer \answerNA{} means that the paper does not involve crowdsourcing nor research with human subjects.
        \item Including this information in the supplemental material is fine, but if the main contribution of the paper involves human subjects, then as much detail as possible should be included in the main paper. 
        \item According to the NeurIPS Code of Ethics, workers involved in data collection, curation, or other labor should be paid at least the minimum wage in the country of the data collector. 
    \end{itemize}

\item {\bf Institutional review board (IRB) approvals or equivalent for research with human subjects}
    \item[] Question: Does the paper describe potential risks incurred by study participants, whether such risks were disclosed to the subjects, and whether Institutional Review Board (IRB) approvals (or an equivalent approval/review based on the requirements of your country or institution) were obtained?
    \item[] Answer: \answerNA{} 
    \item[] Justification: Our experiment is conducted in a simulation without human subjects. So, our research does not need for IRB approval.
    \item[] Guidelines:
    \begin{itemize}
        \item The answer \answerNA{} means that the paper does not involve crowdsourcing nor research with human subjects.
        \item Depending on the country in which research is conducted, IRB approval (or equivalent) may be required for any human subjects research. If you obtained IRB approval, you should clearly state this in the paper. 
        \item We recognize that the procedures for this may vary significantly between institutions and locations, and we expect authors to adhere to the NeurIPS Code of Ethics and the guidelines for their institution. 
        \item For initial submissions, do not include any information that would break anonymity (if applicable), such as the institution conducting the review.
    \end{itemize}

\item {\bf Declaration of LLM usage}
    \item[] Question: Does the paper describe the usage of LLMs if it is an important, original, or non-standard component of the core methods in this research? Note that if the LLM is used only for writing, editing, or formatting purposes and does \emph{not} impact the core methodology, scientific rigor, or originality of the research, declaration is not required.
    \item[] Answer: \answerNA{} 
    \item[] Justification:
    \item[] Guidelines:
    \begin{itemize}
        \item The answer \answerNA{} means that the core method development in this research does not involve LLMs as any important, original, or non-standard components.
        \item Please refer to our LLM policy in the NeurIPS handbook for what should or should not be described.
    \end{itemize}

\end{enumerate}